\documentclass{article}

\usepackage[nonatbib,final]{neurips_2018}

\usepackage[utf8]{inputenc} 
\usepackage[T1]{fontenc}    
\usepackage{hyperref}       
\usepackage{url}            
\usepackage{booktabs}       
\usepackage{amsfonts}       
\usepackage{nicefrac}       
\usepackage{microtype}      

\usepackage{microtype}
\usepackage{booktabs}
\usepackage{times}
\usepackage{graphicx} 
\usepackage{subfigure}
\usepackage{authblk}
\usepackage{algorithm}
\usepackage{algorithmic}

\usepackage{amsmath,amssymb,amsthm}

\usepackage{caption}
\usepackage{tikz}
\usepackage{xcolor}
\usetikzlibrary{arrows}
\usepackage{threeparttable}
\usepackage{multirow}
\usepackage{cancel}
\usepackage{tabularx}
\usepackage{textcomp}
\usepackage{threeparttable}

\usepackage{paralist}
\usepackage{todonotes}

\newcommand{\nn}{\nonumber}
\newcommand{\defeq}{\triangleq}
\newcommand{\mc}{\mathcal}

\newcommand{\argmax}{\mathrm{argmax}}

\newcommand{\modelname}{M-Walk}

\title{\modelname: Learning to Walk over Graphs\\ using Monte Carlo Tree Search}

\author{
\thanks{Yelong Shen, Jianshu Chen, and Po-Sen Huang contributed equally to the paper. The work was done when Yelong Shen and Jianshu Chen were with Microsoft Research. $^\star$Po-Sen Huang is now at DeepMind (Email: \texttt{posenhuang@google.com}).}~~{\bf Yelong Shen}$^1$,
$^*${\bf Jianshu Chen}$^1$, 
$^*${\bf Po-Sen Huang}$^2$$^\star$,
{\bf Yuqing Guo}$^2$,
{\bf and Jianfeng Gao}$^2$ \\
$^1$Tencent AI Lab, Bellevue, WA, USA. \\
\texttt{\{yelongshen, jianshuchen\}@tencent.com}  \\
$^2$Microsoft Research, Redmond, WA, USA \\
\texttt{\{yuqguo, jfgao\}@microsoft.com}
}

\begin{document}

\maketitle
		
\begin{abstract}
Learning to walk over a graph towards a target node for a given query and a source node is an important problem in applications such as knowledge base completion (KBC). It can be formulated as a reinforcement learning (RL) problem with a known state transition model. To overcome the challenge of sparse rewards, we develop a graph-walking agent called \modelname, which consists of a deep recurrent neural network (RNN) and Monte Carlo Tree Search (MCTS). The RNN encodes the state (i.e., history of the walked path) and maps it separately to a policy and Q-values. In order to effectively train the agent from sparse rewards, we combine MCTS with the neural policy to generate trajectories yielding more positive rewards. From these trajectories, the network is improved in an off-policy manner using Q-learning, which modifies the RNN policy via parameter sharing. Our proposed RL algorithm repeatedly applies this policy-improvement step to learn the model. At test time, MCTS is combined with the neural policy to predict the target node. Experimental results on several graph-walking benchmarks show that \modelname~is able to learn better policies than other RL-based methods, which are mainly based on policy gradients. M-Walk also outperforms traditional KBC baselines.
\end{abstract}

	\section{Introduction}
	
\begin{figure}[t!]
    \centerline{
        \subfigure[{\small An example of Knowledge Base Completion}]{
            \includegraphics[width=0.42\textwidth]{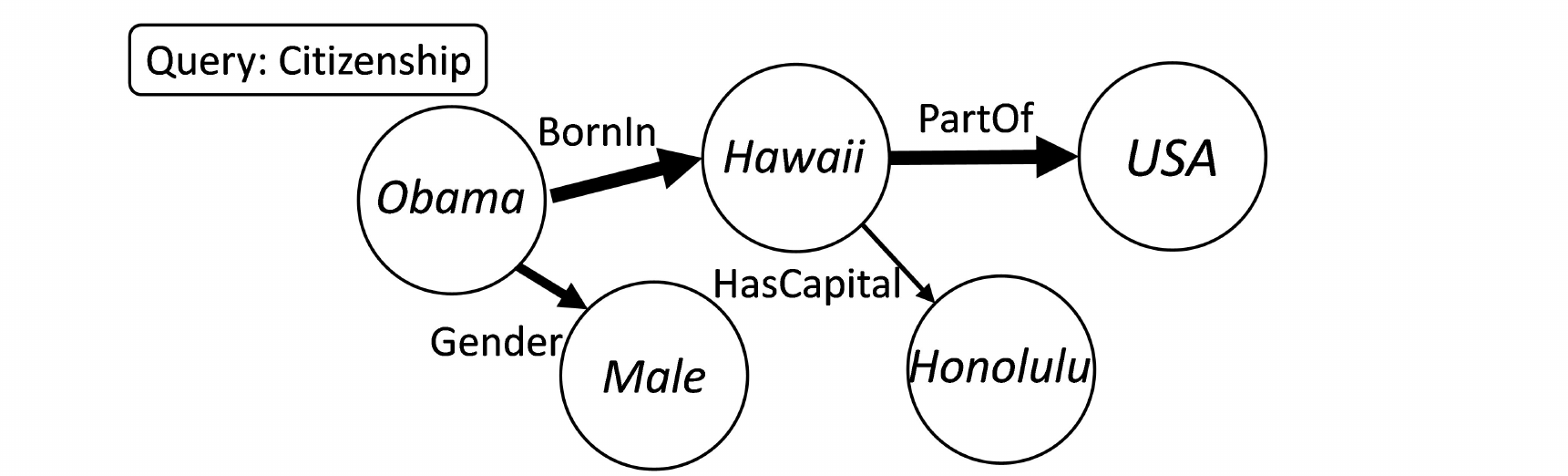}
            \label{fig:graph_examples}
        }
        \subfigure[{\small The corresponding Markov Decision Process}]{
            \includegraphics[width=0.5\textwidth]{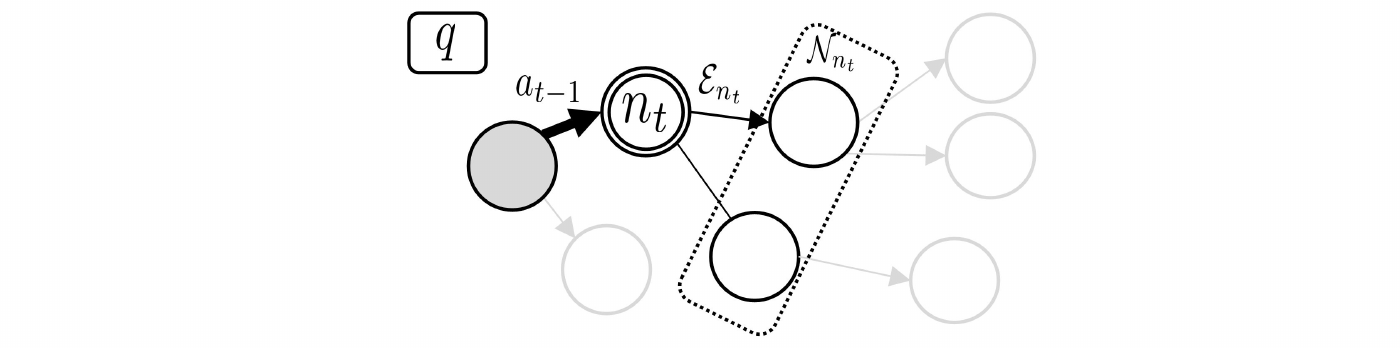}
            \label{fig:state_transition}
        }
    }
    \caption{{\small An example of Knowledge Base Completion and its formulation as a Markov Decision Process. (a) We want to identify the target node $n_T$ = \texttt{USA} for a given pair of query $q$ = \texttt{Citizenship} and source node $n_S$ = \texttt{Obama}. (b) The activated circles and edges (in black lines) denote all the observed information up to time $t$ (i.e., the state $s_t$). The double circle denotes the current node $n_t$, while $\mc{E}_{n_t}$ and $\mc{N}_{n_t}$ denote the edges and nodes connected to the current node.}}
    \end{figure}

	We consider the problem of learning to walk over a graph in order to find a target node for a given source node and a query. Such problems appear in, for example, knowledge base completion (KBC) \cite{DeepPath,lin2015learning,trouillon2017knowledge,NickelTrKr11,gaosurvey}. A knowledge graph is a structured representation of world knowledge in the form of entities and their relations (e.g., Figure \ref{fig:graph_examples}), and has a wide range of downstream applications such as question answering. Although a typical knowledge graph may contain millions of entities and billions of relations, it is usually far from complete. KBC aims to predict the missing relations between entities using information from the existing knowledge graph.
	More formally, let $\mc{G} = (\mc{N},\mc{E})$ denote a graph, which consists of a set of nodes, $\mc{N}=\{n_{i}\}$, and a set of edges, $\mc{E}=\{e_{ij}\}$, that connect the nodes, and let $q$ denote an input query. The problem is stated as using the graph $\mc{G}$, the source node $n_S \in \mc{N}$ and the query $q$ as inputs to predict the target node $n_T \in \mc{N}$. In KBC tasks, $\mc{G}$ is a given knowledge graph, $\mc{N}$ is a collection of entities (nodes), and $\mc{E}$ is a set of relations (edges) that connect the entities. In the example in Figure \ref{fig:graph_examples}, the objective of KBC is to identify the target node $n_T$ = \texttt{USA} for the given head entity $n_S$ = \texttt{Obama} and the given query $q$ = \textsc{Citizenship}.

	The problem can also be understood as constructing a function $f(\mc{G}, n_S, q)$ to predict $n_T$, where the functional form of $f(\cdot)$ is generally unknown and has to be learned from a training dataset consisting of samples like $(n_S, q, n_T)$. In this work, we model $f(\mc{G}, n_S, q)$ by means of a graph-walking \emph{agent} that intelligently navigates through a \emph{subset} of nodes in the graph from $n_S$ towards $n_T$. Since $n_T$ is unknown, the problem cannot be solved by conventional search algorithms such as $A^*$-search \cite{hart1968formal}, which seeks to find paths between the given source and target nodes. Instead, the \emph{agent} needs to \emph{learn} its search policy from the training dataset so that, after training is complete, the agent knows how to walk over the graph to reach the correct target node $n_T$ for an \emph{unseen} pair of $(n_S, q)$. Moreover, each training sample is in the form of ``(source node, query, target node)'', and there is no intermediate supervision for the correct search path. Instead, the agent receives only delayed \emph{evaluative} feedback: when the agent correctly (or incorrectly) predicts the target node in the training set, the agent will receive a positive (or zero) reward. For this reason, we formulate the problem as a Markov decision process (MDP) and train the agent by reinforcement learning (RL) \cite{sutton1998reinforcement}. 
	
	The problem poses two major challenges. Firstly, since the state of the MDP is the entire trajectory, reaching a correct decision usually requires not just the query, but also the entire history of traversed nodes. For the KBC example in Figure \ref{fig:graph_examples}, having access to the current node $n_t$ = \texttt{Hawaii} alone is not sufficient to know that the best action is moving to $n_{t+1}$ = \texttt{USA}. Instead, the agent must track the entire history, including the input query $q$ = \texttt{Citizenship}, to reach this decision. Secondly, the reward is sparse, being received only at the end of a search path, for instance, after correctly predicting $n_T$=\texttt{USA}. 
	
	In this paper, we develop a neural graph-walking agent, named \emph{\modelname}, that effectively addresses these two challenges. First, M-Walk uses a novel recurrent neural network (RNN) architecture to encode the entire history of the trajectory into a vector representation, which is further used to model the policy and the Q-function. Second, to address the challenge of sparse rewards, M-Walk exploits the fact that the MDP transition model is known and deterministic.\footnote{Whenever the agent takes an action, by selecting an edge connected to a next node, the identity of the next node (which the environment will transition to) is already known. Details can be found in Section \ref{Sec:MDPformulation}.} Specifically, it combines Monte Carlo Tree Search (MCTS) with the RNN to generate trajectories that obtain significantly more positive rewards than using the RNN policy alone. These trajectories can be viewed as being generated from an improved version of the RNN policy. But while these trajectories can improve the RNN policy, their off-policy nature prevents them from being leveraged by policy gradient RL methods. To solve this problem, we design a structure for sharing parameters between the Q-value network and the RNN's policy network. This allows the policy network to be indirectly improved through Q-learning over the off-policy trajectories. Our method is in sharp contrast to existing RL-based methods for KBC, which use a policy gradients (REINFORCE) method \cite{williams1992simple} and usually require a large number of rollouts to obtain a trajectory with a positive reward, especially in the early stages of learning \cite{gu2016q,wu2017scalable,kakade2002natural}. Experimental results on several benchmarks, including a synthetic task and several real-world KBC tasks, show that our approach learns better policies than previous RL-based methods and traditional KBC methods.

	The rest of the paper is organized as follows: Section \ref{Sec:\modelname} develops the \modelname~agent, including the model architecture, the training and testing algorithms.\footnote{The code of this paper is available at: \url{https://github.com/yelongshen/GraphWalk}} Experimental results are presented in Section \ref{Sec:Experiments}. Finally, we discuss related work in Section \ref{Sec:RelatedWork} and conclude the paper in Section \ref{Sec:Conclusion}.

	\section{Graph Walking as a Markov Decision Process}
	\label{Sec:MDPformulation}
	
	In this section, we formulate the graph-walking problem as a Markov Decision Process (MDP), which is defined by the tuple $(\mc{S}, \mc{A}, \mc{R}, \mc{P})$, where $\mc{S}$ is the set of states, $\mc{A}$ is the set of actions, $\mc{R}$ is the reward function, and $\mc{P}$ is the state transition probability. We further define $\mc{S}$, $\mc{A}$, $\mc{R}$ and $\mc{P}$ below. Figure \ref{fig:state_transition} illustrates the MDP corresponding to the KBC example of Figure \ref{fig:graph_examples}. Let $s_t \in \mc{S}$ denote the state at time $t$. Recalling that the agent needs the entire history of traversed nodes and the query to make a correct decision, we define $s_t$ by the following recursion:
    	\begin{align}
    	    s_t	    &=	    
    	                s_{t-1} \cup \{a_{t-1}, n_t, \mc{E}_{n_t}, \mc{N}_{n_t}\}, \qquad s_0 \defeq \{q, n_S, \mc{E}_{n_S}, \mc{N}_{n_S}\}
    	\label{Equ:\modelname:s_t1}
    	\end{align}
    \noindent where $a_t \in \mc{A}$ denotes the action selected by the agent at time $t$, $n_t \in \mc{G}$ denotes the currently visited node at time $t$,  $\mc{E}_{n_t} \subset \mc{E}$ is the set of all edges connected to $n_t$, and $\mc{N}_{n_t} \subset \mc{N}$ is the set of all nodes connected to $n_t$ (i.e., the neighborhood). Note that state $s_t$ is a collection of (i) all the traversed nodes (along with their edges and neighborhoods) up to time $t$, (ii) all the previously selected (up to time $t-1$) actions, and (iii) the initial query $q$. The set $\mc{S}$ consists of all the possible values of $\{s_t, t \ge 0\}$. Based on $s_t$, the agent takes one of the following actions at each time $t$: (i) choosing an edge in $\mc{E}_{n_{t}}$ and moving to the next node $n_{t+1} \in \mc{N}_{n_{t}}$, or (ii) terminating the walk (denoted as the ``STOP'' action). Once the STOP action is selected, the MDP reaches the terminal state and outputs $\hat{n}_T = n_{t}$ as a prediction of the target node $n_T$. Therefore, we define the set of feasible actions at time $t$ as $\mc{A}_t \defeq \mc{E}_{n_t} \cup \{ \mathrm{STOP}\}$, which is usually time-varying. The entire action space $\mc{A}$ is the union of all $\mc{A}_t$, i.e., $\mc{A} = \cup_t \mc{A}_t$. Recall that the training set consists of samples in the form of $(n_S, q, n_T)$. The reward is defined to be $+1$ when the predicted target node $\hat{n}_T$ is the same as $n_T$ (i.e., $\hat{n}_T = n_T$), and zero otherwise. In the example of Figure \ref{fig:graph_examples}, for a training sample (\texttt{Obama}, \texttt{Citizenship}, \texttt{USA}), if the agent successfully navigates from \texttt{Obama} to \texttt{USA} and correctly predicts $\hat{n}_T$ = \texttt{USA}, the reward is $+1$. Otherwise, it will be $0$. The rewards are sparse because positive reward can be received only at the end of a correct path. Furthermore, since the graph $\mc{G}$ is known and static, the MDP transition probability $p(s_{t} | s_{t-1}, a_{t-1})$ is \emph{known} and \emph{deterministic}, and is defined by \eqref{Equ:\modelname:s_t1}. To see this, we observe from Figure \ref{fig:state_transition} that once an action $a_t$ (i.e., an edge in $\mc{E}_{n_t}$ or ``STOP'') is selected, the next node $n_{t+1}$ and its associated $\mc{E}_{n_{t+1}}$ and $\mc{N}_{n_{t+1}}$ are known. By \eqref{Equ:\modelname:s_t1} (with $t$ replaced by $t+1$), this means that the next state $s_{t+1}$ is determined. This important (model-based) knowledge will be exploited to overcome the sparse-reward problem using MCTS and significantly improve the performance of our method (see Sections \ref{Sec:\modelname}--\ref{Sec:Experiments} below). 
    
    We further define $\pi_{\theta}(a_t|s_t)$ and $Q_{\theta}(s_t, a_t)$ to be the policy and the Q-function, respectively, where $\theta$ is a set of model parameters. The policy $\pi_{\theta}(a_t|s_t)$ denotes the probability of taking action $a_t$ given the current state $s_t$. In \modelname, it is used as a prior to bias the MCTS search. And $Q_{\theta}(s_t,a_t)$ defines the long-term reward of taking action $a_t$ at state $s_t$ and then following the optimal policy thereafter. The objective is to learn a policy that maximizes the terminal rewards, i.e., correctly identifies the target node with high probability. We now proceed to explain how to model and jointly learn $\pi_{\theta}$ and $Q_{\theta}$ to achieve this objective.

	\section{The \modelname~Agent}
	\label{Sec:\modelname}
	
	In this section, we develop a neural graph-walking agent named \modelname~(i.e., MCTS for graph Walking), which consists of (i) a novel neural architecture for jointly modeling $\pi_{\theta}$ and $Q_{\theta}$, and (ii) Monte Carlo Tree Search (MCTS). We first introduce the overall neural architecture and then explain how MCTS is used during the training and testing stages. Finally, we describe some further details of the neural architecture. Our discussion focuses on addressing the two challenges described earlier: history-dependent state and sparse rewards.
	
	\subsection{The neural architecture for jointly modeling $\pi_{\theta}$ and $Q_{\theta}$}
	\label{Sec:\modelname:Model}
	
	Recall from Section \ref{Sec:MDPformulation} (e.g., \eqref{Equ:\modelname:s_t1}) that one challenge in applying RL to the graph-walking problem is that the state $s_t$ nominally includes the entire history of observations. To address this problem, we propose a special RNN encoding the state $s_t$ at each time $t$ into a vector representation, $h_t = \mathrm{ENC}_{\theta_e}(s_t)$, where $\theta_e$ is the associated model parameter. We defer the discussion of this RNN state encoder to Section \ref{Sec:\modelname:RNNStateEncoder}, and focus in this section on how to use $h_t$ to jointly model $\pi_{\theta}$ and $Q_{\theta}$. Specifically, the vector  $h_t$ consists of several sub-vectors of the same dimension $M$: $h_{S,t}$,  $\{h_{n',t}: n' \in \mc{N}_{n_t}\}$ and $h_{A,t}$. Each sub-vector encodes part of the state $s_t$ in \eqref{Equ:\modelname:s_t1}. For instance, the vector $h_{S,t}$ encodes $(s_{t-1}, a_{t-1}, n_t)$, which characterizes the history in the state. The vector $h_{n',t}$ encodes the (neighboring) node $n'$ and the edge $e_{n_t,n'}$ connected to $n_t$, which can be viewed as a vector representation of the $n'$-th candidate action (excluding the STOP action). And the vector $h_{A,t}$ is a vector summarization of $\mc{E}_{n_t}$ and $\mc{N}_{n_t}$, which is used to model the STOP action probability. In summary, we use the sub-vectors to model $\pi_{\theta}$ and $Q_{\theta}$ according to:
    	\begin{align}
        	u_0		&=
        	f_{\theta_{\pi}}(h_{S,t}, h_{A,t}), \quad
        	u_{n'}	=
        	\langle h_{S,t}, h_{n',t} \rangle, \quad n' \in \mc{N}_{n_t}
        	\label{Equ:\modelname:u}
        	\\
        	Q_{\theta}(s_t, \cdot)
        	&=
        	\sigma(u_0, u_{n_1'},\ldots,u_{n_k'}),
        	\;
        	\pi_{\theta}(\cdot | s_t)
        	=
        	\phi_{\tau}(u_0, u_{n_1'},\ldots,u_{n_k'})
        	\label{Equ:\modelname:PolicyNet}
    	\end{align}
	where $\langle \cdot, \cdot \rangle$ denotes inner product, $f_{\theta_\pi}(\cdot)$ is a fully-connected neural network with model parameter $\theta_\pi$, $\sigma(\cdot)$ denotes the element-wise sigmoid function, and $\phi_{\tau}(\cdot)$ is the softmax function with temperature parameter $\tau$. Note that we use the inner product between the vectors $h_{S,t}$ and $h_{n',t}$ to compute the (pre-softmax) score $u_{n'}$ for choosing the $n'$-th candidate action, where $n' \in \mc{N}_{n_t}$. The inner product operation has been shown to be useful in modeling Q-functions when the candidate actions are described by vector representations \cite{he2015deep,chen2017q} and in solving other problems \cite{vinyals2015pointer,bello2016neural}. Moreover, the value of $u_0$ is computed by $f_{\theta_\pi}(\cdot)$ using $h_{S,t}$ and $h_{A,t}$, where $u_0$ gives the (pre-softmax) score for choosing the STOP action. We model the Q-function by applying element-wise sigmoid to $u_0,u_{n_1'},\ldots,u_{n_k'}$, and we model the policy by applying the softmax operation to the \emph{same} set of $u_0,u_{n_1'},\ldots,u_{n_k'}$.\footnote{An alternative choice is applying softmax to the Q-function to get the policy, which is known as softmax selection \cite{sutton1998reinforcement}. We found in our experiments that these two designs do not differ much in performance.} Note that the policy network and the Q-network share the same set of model parameters. We will explain in Section \ref{Sec:\modelname:Training} how such parameter sharing enables indirect updates to the policy $\pi_{\theta}$ via Q-learning from off-policy data.

	\begin{figure}[t!]
		\centering
		\subfigure[]{
			\label{Fig:NN:PolicyValueQ}
			\includegraphics[width=0.42\textwidth]{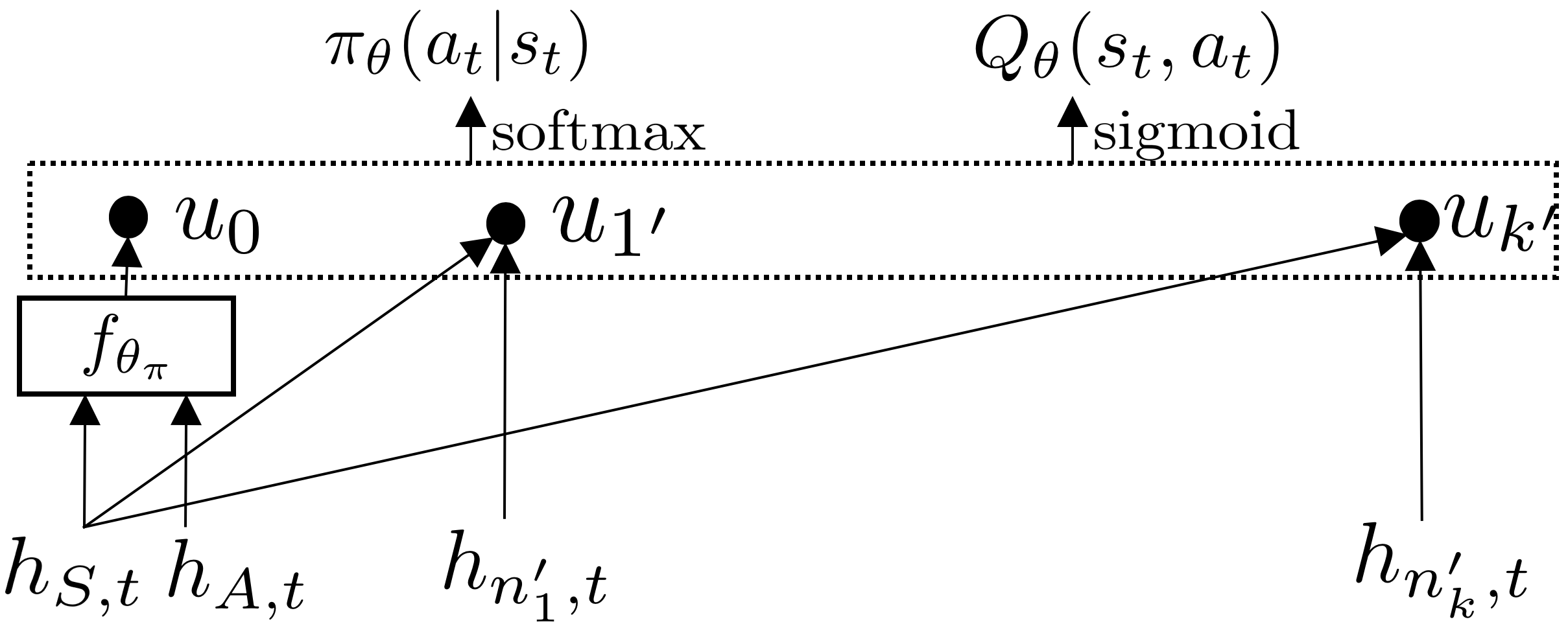}
		}\hfil
		\subfigure[]{
			\label{Fig:NN:RNNStateEncoder}
			\includegraphics[width=0.4\textwidth]{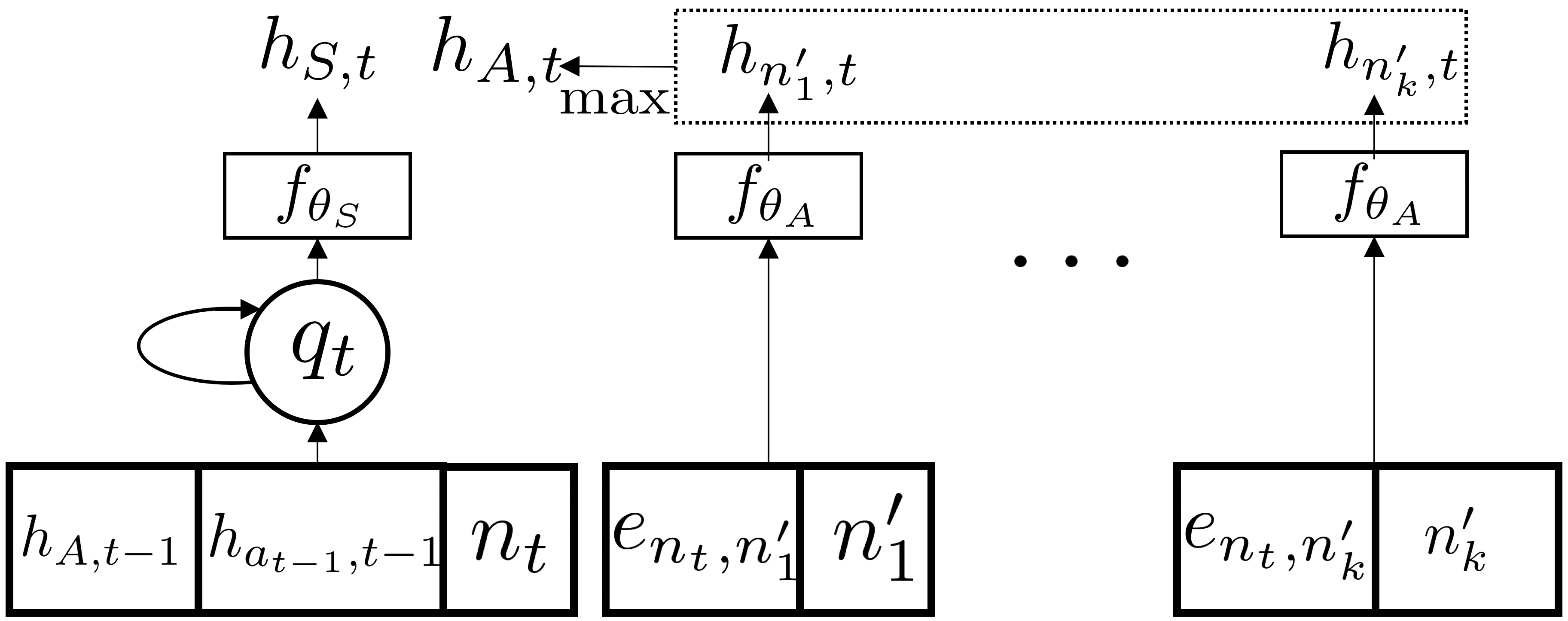}
		}
		\caption{{\small The neural architecture for \modelname. (a) The vector representation of the state is mapped into $\pi_{\theta}$ and $Q_{\theta}$. (b) The GRU-RNN state encoder maps the state into its vector representation $h_t$. Note that the inputs $h_{A,t-1}$ and $h_{a_{t-1},t-1}$ are from the output of the previous time step $t-1$.}}
		\label{fig:DeepRNN}
	\end{figure}

	\subsection{The training algorithm}
	\label{Sec:\modelname:Training}

\begin{figure}[t!]
    \centerline{
        \subfigure[\scriptsize An example of MCTS path (in red) in \modelname]{
    	\includegraphics[width=0.4\textwidth]{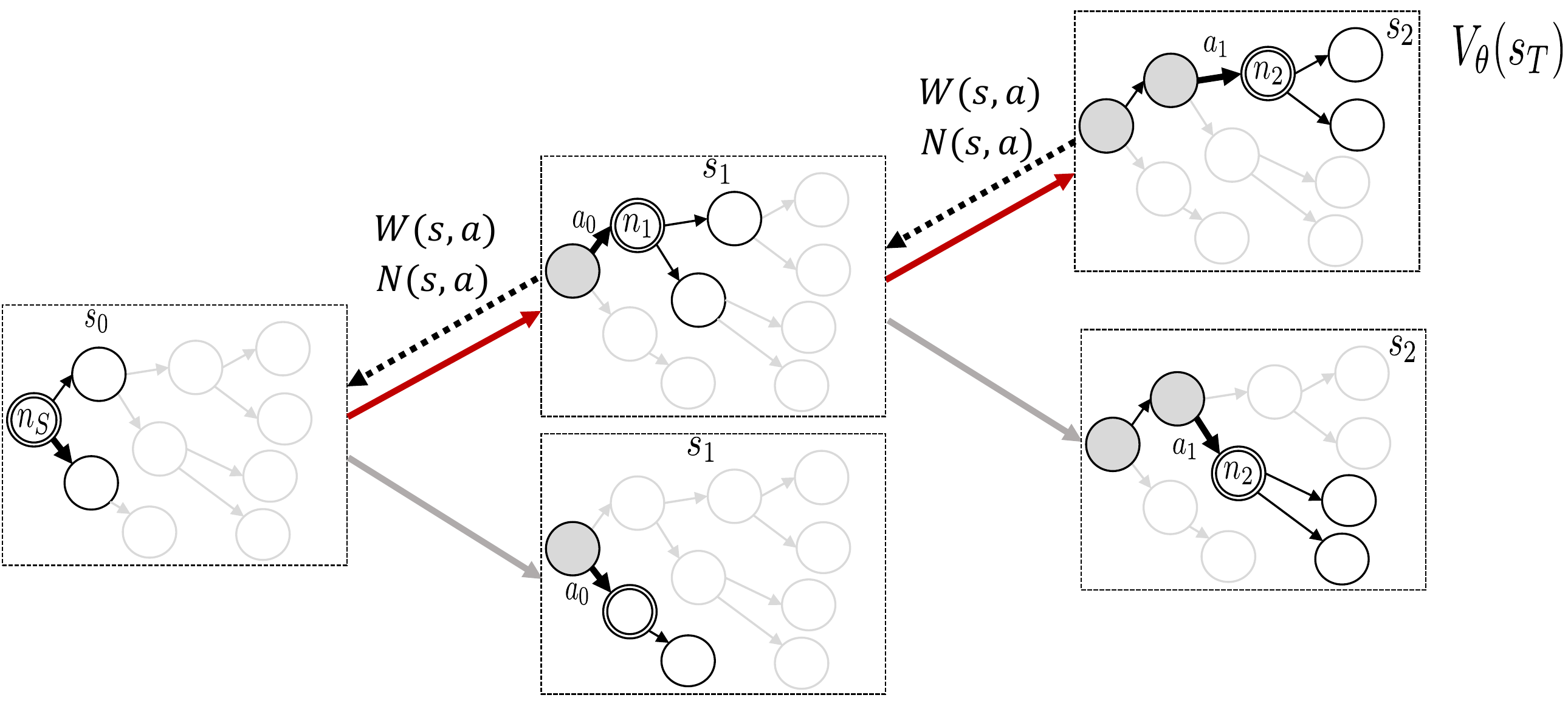}
		\label{fig:mcts}
        }
        \hfil
        \subfigure[\scriptsize Iterative policy improvement in M-Walk]{
            \includegraphics[width=0.45\textwidth]{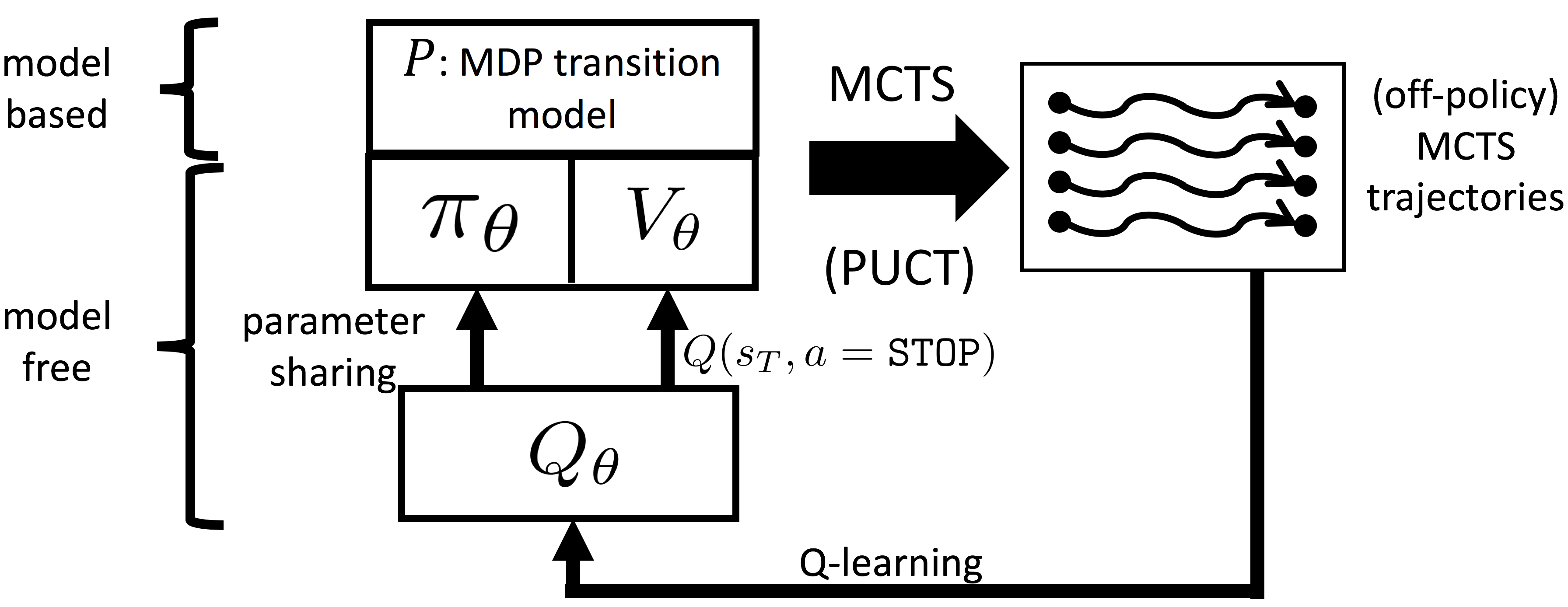}
            \label{fig:framework_idea}
        }
    }
    \caption{{\small MCTS is used to generate trajectories for iterative policy improvement in \modelname.}}
    \end{figure}

	We now discuss how to train the model parameters $\theta$ (including $\theta_{\pi}$ and $\theta_e$) from a training dataset $\{(n_S,q,n_T)\}$ using reinforcement learning. One approach is the policy gradient method (REINFORCE) \cite{williams1992simple,sutton2000policy}, which uses the current policy $\pi_{\theta}(a_t|s_t)$ to roll out multiple trajectories $(s_0, a_0, r_0, s_1,\ldots)$ to estimate a stochastic gradient, and then updates the policy $\pi_{\theta}$ via stochastic gradient ascent. Previous RL-based KBC methods \cite{DeepPath,GoforaWalk} typically use REINFORCE to learn the policy. However, policy gradient methods generally suffer from low sample efficiency, especially when the reward signal is sparse, because large numbers of Monte Carlo rollouts are usually needed to obtain many trajectories with positive terminal reward, particularly in the early stages of learning. To address this challenge, we develop a novel RL algorithm that uses MCTS to exploit the deterministic MDP transition defined in \eqref{Equ:\modelname:s_t1}. Specifically, on each MCTS simulation, a trajectory is rolled out by selecting actions according to a variant of the PUCT algorithm \cite{Rosin2011, silver2017mastering} from the root state $s_0$ (defined in \eqref{Equ:\modelname:s_t1}):
    	\begin{align}
        	a_t     =
        	            \argmax_a
        	            \Big\{ 
        	                c \cdot \pi_{\theta}(a | s_t)^{\beta}
        	                \sqrt{
        	                    \textstyle \sum_{a'} N(s_t, a')
        	                    }
        	                \big/
        	                (1 \!+\! N(s_t, a)) 
        	                \!+\! 
        	                W(s_t, a)/N(s_t, a)
        	            \Big\}	   	
    	\label{Equ:\modelname:PUCT}
    	\end{align}
	where $\pi_{\theta}(a|s)$ is the policy defined in Section \ref{Sec:\modelname:Model}, $c$ and $\beta$ are two constants that control the level of exploration, and $N(s,a)$ and $W(s,a)$ are the visit count and the total action reward accumulated on the $(s,a)$-th edge on the MCTS tree. Overall, PUCT treats $\pi_\theta$ as a prior probability to bias the MCTS search; PUCT initially prefers actions with high values of $\pi_{\theta}$ and low visit count $N(s,a)$ (because the first term in \eqref{Equ:\modelname:PUCT} is large), but then asympotically prefers actions with high value (because the first term in \eqref{Equ:\modelname:PUCT} vanishes and the second term $W(s,a)/N(s,a)$ dominates). When PUCT selects the STOP action or the maximum search horizon has been reached, MCTS completes one simulation and updates $W(s,a)$ and $N(s,a)$ using $V_{\theta}(s_T) = Q_{\theta}(s_T, a=\tt{STOP})$. (See Figure \ref{fig:mcts} for an example and Appendix \ref{Appendix:AlgDetails:MCTS} for more details.) The key idea of our method is that running multiple MCTS simulations generates a set of trajectories with more positive rewards (see Section \ref{Sec:Experiments} for more analysis), which can also be viewed as being generated by an improved policy $\pi_{\theta}$. Therefore, learning from these trajectories can further improve $\pi_{\theta}$. Our RL algorithm repeatedly applies this policy-improvement step to refine the policy. However, since these trajectories are generated by a policy that is different from $\pi_{\theta}$, they are off-policy data, breaking the assumptions inherent in policy gradient methods. For this reason, we instead update the Q-network from these trajectories in an off-policy manner using Q-learning: $\theta \leftarrow \theta  + \alpha \cdot \nabla_{\theta} Q_{\theta}(s_t,a_t) \times ( r(s_t,a_t) + \gamma \max_{a'} Q_{\theta}(s_{t+1},a') - Q_{\theta}(s_t,a_t) )$. Recall from Section \ref{Sec:\modelname:Model} that $\pi_{\theta}$ and $Q_{\theta}(s,a)$ share the same set of model parameters; once the Q-network is updated, the policy network $\pi_{\theta}$ will also be automatically improved. Finally, the new $\pi_{\theta}$ is used to control the MCTS in the next iteration. The main idea of the training algorithm is summarized in Figure \ref{fig:framework_idea}.

	\subsection{The prediction algorithm}
	\label{Sec:\modelname:Test}
	
	At test time, we want to infer the target node $n_T$ for an unseen pair of $(n_S, q)$. One approach is to use the learned policy $\pi_{\theta}$ to walk through the graph $\mc{G}$ to find $n_T$. However, this would not exploit the known MDP transition model \eqref{Equ:\modelname:s_t1}. Instead, we combine the learned $\pi_{\theta}$ and $Q_{\theta}$ with MCTS to generate an MCTS search tree, as in the training stage. Note that there could be multiple paths that reach the same terminal node $n \in \mc{G}$, meaning that there could be multiple leaf states in MCTS corresponding to that node. Therefore, the prediction results from these MCTS leaf states need to be merged into one score to rank the node $n$. Specifically, we use
	$\mathrm{Score}(n) = \textstyle \sum_{ s_T \rightarrow n} N(s_T, a_T)/{N} \times Q_{\theta}(s_T,\tt{STOP})$,    
    where $N$ is the total number of MCTS simulations, and the summation is over all the leaf states $s_T$ that correspond to the same node $n \in \mc{G}$. 
    $\mathrm{Score}(n)$ is a weighted average of the terminal state values associated with the same candidate node $n$.\footnote{There could be alternative ways to compute the score, such as $\mathrm{Score}(n) = \max_{s_T \rightarrow n} Q_{\theta}(s_T,\tt{STOP})$. However, we found in our (unreported) experiments that they do not make much difference.} Among all the candidates nodes, we select the predicted target node to be the one with the highest score: $\hat{n}_T = \argmax_{n} \mathrm{Score}(n)$.

	\subsection{The RNN state encoder}
	\label{Sec:\modelname:RNNStateEncoder}
	
	We now discuss the details of the RNN state encoder $h_t = \mathrm{ENC}_{\theta_e}(s_t)$, where $\theta_e \defeq \{\theta_A,\theta_S,\theta_q\}$, as shown in Figure \ref{Fig:NN:RNNStateEncoder}. Specifically, we explain how the sub-vectors of $h_t$ are computed. We introduce $q_t \defeq s_{t-1} \cup \{a_{t-1}, n_t\}$ as an auxiliary variable. Then, the state $s_t$ in \eqref{Equ:\modelname:s_t1} can be written as $s_t = q_t \cup \{\mc{E}_{n_t}, \mc{N}_{n_t}\}$. Note that the state $s_t$ is composed of two parts: (i) $\mc{E}_{n_t}$ and $\mc{N}_{n_t}$, which represent the candidate actions to be selected (excluding the STOP action), and (ii) $q_t$, which represents the history. We use two different neural networks to encode these separately. For the $n'$-th candidate action ($n' \in \mc{N}_{n_t}$), we concatenate $n'$ with its associated $e_{n_t,n'} \in \mc{E}_{n_t}$ and input them into a fully connected network (FCN) $f_{\theta_A}(\cdot)$ to compute their joint vector representation $h_{n',t}$, where $\theta_A$ is the model parameter. Recall that the action space $\mc{A}_t = \mc{E}_{n_t} \cup \{\mathrm{STOP}\}$ can be time-varying when the size of $\mc{E}_{n_t}$ changes over time. To address this issue, we apply the same FCN $f_{\theta}(\cdot)$ to different $(n',e_{n_t,n'})$ to obtain their respective representations. Then, we use a coordinate-wise max-pooling operation over $\{h_{n',t}: n' \in \mc{N}_{n_t}\}$ to obtain a (fixed-length) overall vector representation of $\{\mc{E}_{n_t}, \mc{N}_{n_t}\}$. To encode $q_t$, we call upon the following recursion for $q_t$ (see Appendix \ref{Appendix:qt_derivation} for the derivation): $q_{t+1} = q_t \cup \{ \mc{E}_{n_t}, \mc{N}_{n_t}, a_t, n_{t+1} \}$. Inspired by this recursion, we propose using the GRU-RNN \cite{cho2014learning} to encode $q_t$ into a vector representation\footnote{For simplicity, we use the same notation $q_t$ to denote its vector representation.}: $q_{t+1}=f_{\theta_q}(q_{t}, [h_{A,t}, h_{a_{t},t}, n_{t+1}])$ with initialization $q_0=f_{\theta_q}(q,[0,0,n_S])$, where $\theta_q$ is the model parameter, and $h_{a_t,t}$ denotes the vector $h_{n',t}$ at $n'=a_t$. We use $h_{A,t}$ and $h_{a_t,t}$ computed by the FCNs to represent $(\mc{E}_{n_t},\mc{N}_{n_t})$ and $a_t$, respectively. Then, we map $q_t$ to $h_{S,t}$ using another FCN $f_{\theta_S}(\cdot)$. 

	\section{Experiments}
	\label{Sec:Experiments}

	We evaluate and analyze the effectiveness of M-Walk on a synthetic Three Glass Puzzle task and two real-world KBC tasks. We briefly describe the tasks here, and give the experiment details and hyperparameters in Appendix \ref{Appendix:AlgDetails}. 
	
	\paragraph{Three Glass Puzzle} The Three Glass Puzzle \cite{ore1990graphs} is a problem studied in math puzzles and graph theory. It involves three milk containers $\mathcal{A}$, $\mathcal{B}$, and $\mathcal{C}$, with capacities $A$, $B$ and $C$ liters, respectively. The containers display no intermediate markings. There are three feasible actions at each time step: (i) \emph{fill} a container (to its capacity),  (ii) \emph{empty} all of its liquid, and (iii) \emph{pour} its liquid into another container (up to its capacity). The objective of the problem is, given a desired volume $q$, to take a sequence of actions on the three containers after which one of them contains $q$ liters of liquid. We formulate this as a graph-walking problem; in the graph $\mc{G}$, each node $n=(a,b,c)$ denotes the amounts of remaining liquid in the three containers, each edge denotes one of the three feasible actions, and the input query is the desired volume $q$. The reward is $+1$ when the agent successfully fills one of the containers to $q$ and $0$ otherwise (see Appendix \ref{Appendix:puzzle_details} for the details).
	We use vanilla policy gradient (REINFORCE) \cite{williams1992simple} as the baseline, with task success rate as the evaluation metric.
	
	\paragraph{Knowledge Base Completion}
    
	We use WN18RR and NELL995 knowledge graph datasets for evaluation. WN18RR \cite{dettmers2018conve} is created from the original WN18 \cite{bordes2013translating} by removing various sources of test leakage, making the dataset more challenging. The NELL995 dataset was released by \cite{DeepPath} and has separate graphs for each query relation.
	We use the same data split and preprocessing protocol as in \cite{dettmers2018conve} for WN18RR and in \cite{DeepPath, GoforaWalk} for NELL995. 
	As in \cite{DeepPath, GoforaWalk}, we study the 10 relation tasks of NELL995 separately.
	We use HITS@1,3 and mean reciprocal rank (MRR) as the evaluation metrics for WN18RR, and use mean average precision (MAP) for NELL995,\footnote{We use these metrics in order to be consistent with \cite{DeepPath,GoforaWalk}. We also report the HITS and MRR scores for NELL995 in Table \ref{tab:nell995_hits} of the supplementary material.} where HITS@$K$ computes the percentage of the desired entities being ranked among the top-$K$ list, and MRR computes an average of the reciprocal rank of the desired entities. 
	We compare against RL-based methods \cite{DeepPath, GoforaWalk}, embedding-based models (including DistMult \cite{DBLP:journals/corr/YangYHGD14a}, ComplEx \cite{trouillon2016complex} and ConvE \cite{dettmers2018conve}) and recent work in logical rules (NeuralLP) \cite{yang2017differentiable}. 
	For all the baseline methods, we used the implementation released by the corresponding authors with their best-reported hyperparameter settings.\footnote{ConvE: https://github.com/TimDettmers/ConvE, Neural-LP: https://github.com/fanyangxyz/Neural-LP/, DeepPath: https://github.com/xwhan/DeepPath, MINERVA: https://github.com/shehzaadzd/MINERVA/} The details of the hyperparameters for \modelname~are described in Appendix \ref{Appendix:kbc_details} of the supplementary material.

	\subsection{Performance of \modelname}

	We first report the overall performance of the \modelname~algorithm on the three tasks and compare it with other baseline methods. We ran the experiments three times and report the means and standard deviations (except for PRA, TransE, and TransR on NELL995, whose results are directly quoted from \cite{DeepPath}). On the Three Glass Puzzle task, \modelname~significantly outperforms the baseline: the best model of \modelname~achieves an accuracy of $(99.0 \pm 1.0)\%$ while the best REINFORCE method achieves $(49.0 \pm 2.6)\%$ (see Appendix \ref{Appendix:AdditionalExperiments} for more experiments with different settings on this task). For the two KBC tasks, we report their results in Tables \ref{tab:kbc_results}-\ref{tab:wn18rr_fb15k237}, where PG-Walk and Q-Walk are two methods we created \emph{just for the ablation study in the next section}. The proposed method outperforms previous works in most of the metrics on NELL995 and WN18RR datasets. Additional experiments on the FB15k-237 dataset can be found in Appendix \ref{Appendix:kbc} of the supplementary material.

	\begin{table}[t]
		\begin{center}
        {\scriptsize
		\caption{{\small The MAP scores (\%) on NELL995 task, where we report RL-based methods in terms of ``mean (standard deviation)''. PG-Walk and Q-Walk are methods we created just for the ablation study.}}
		\label{tab:kbc_results}
		 \resizebox{.98\columnwidth}{!}{%
		\begin{tabular}{cccc|ccccc}
		\hline
        Tasks                & \modelname & PG-Walk    & Q-Walk     & MINERVA    & DeepPath    & PRA  & TransE & TransR \\\hline
        AthletePlaysForTeam  & \textbf{84.7} (1.3)       & 80.8 (0.9) & 82.6 (1.2) & 82.7 (0.8) & 72.1  (1.2) & 54.7 & 62.7   & 67.3   \\
        AthletePlaysInLeague & \textbf{97.8} (0.2)       & 96.0 (0.6) & 96.2 (0.8) & 95.2 (0.8) & 92.7  (5.3) & 84.1 & 77.3   & 91.2   \\
        AthleteHomeStadium   & 91.9 (0.1)                & 91.9 (0.3) & 91.1 (1.3) & \textbf{92.8} (0.1) & 84.6  (0.8) & 85.9 & 71.8   & 72.2   \\
        AthletePlaysSport    & 98.3 (0.1)                & 98.0 (0.8) & 97.0 (0.2) & \textbf{98.6} (0.1) & 91.7  (4.1) & 47.4 & 87.6   & 96.3   \\
        TeamPlaySports       & \textbf{88.4} (1.8)       & 87.4 (0.9) & 78.5 (0.6) & 87.5 (0.5) & 69.6  (6.7) & 79.1 & 76.1   & 81.4   \\
        OrgHeadquaterCity    & \textbf{95.0} (0.7)       & 94.0 (0.4) & 94.0 (0.6) & 94.5 (0.3) & 79.0 (0.0)  & 81.1 & 62.0   & 65.7   \\
        WorksFor             & \textbf{84.2} (0.6)       & 84.0 (1.6) & 82.7 (0.2) & 82.7 (0.5) & 69.9  (0.3) & 68.1 & 67.7   & 69.2   \\
        BornLocation         & 81.2 (0.0)                & \textbf{82.3} (0.6) & 81.4 (0.5) & 78.2 (0.0) & 75.5  (0.5) & 66.8 & 71.2   & 81.2   \\
        PersonLeadsOrg       & \textbf{88.8} (0.5)       & 87.2 (0.5) & 86.9 (0.5) & 83.0 (2.6) & 79.0 (1.0)  & 70.0 & 75.1   & 77.2   \\
        OrgHiredPerson       & \textbf{88.8} (0.6)       & 87.2 (0.4) & 87.8 (0.9) & 87.0 (0.3) & 73.8  (1.9) & 59.9 & 71.9   & 73.7  \\
        \hline
         Overall & \textbf{89.9} & 88.9 &     87.8 & 87.6 & 78.8 & 69.7 & 72.3 & 77.5 \\ 
				\hline
			\end{tabular}%
			}%
					}
		\end{center}%
	\end{table}

	
    \begin{table}[t]
    \begin{center}
    \caption{{\small The results on the WN18RR dataset, in the form of ``mean (standard deviation)''.}}
    \label{tab:wn18rr_fb15k237}
    {\scriptsize
		 \resizebox{.98\columnwidth}{!}{%
    \begin{tabular}{lccc|ccccc}
    \hline
    Metric (\%) & \modelname & \multicolumn{1}{c}{PG-Walk} & \multicolumn{1}{c|}{Q-Walk} & MINERVA    & ComplEx    & \multicolumn{1}{l}{ConvE} & \multicolumn{1}{l}{DistMult} & NeuralLP   \\\hline
    HITS@1      & \textbf{41.4 (0.1)}  & 39.3 (0.2)  & 38.2 (0.3)   & 35.1 (0.1) & 38.5 (0.3) & 39.6 (0.3)    & 38.4 (0.4)    & 37.2 (0.1) \\
    HITS@3      & 44.5 (0.2)        & 41.9 (0.1)     & 40.8 (0.4)   & 44.5 (0.4) & 43.9 (0.3) & \textbf{44.7 (0.2)}       & 42.4 (0.3)                   & 43.4 (0.1) \\
    MRR         & \textbf{43.7 (0.1)}       & 41.3 (0.1)                  & 40.1 (0.3)                 & 40.9 (0.1) & 42.2 (0.2) & 43.3 (0.2)   & 41.3 (0.3)   & 43.5 (0.1)
\\ \hline
    \end{tabular}}
    }%
    \end{center}
    \end{table}

	\subsection{Analysis of \modelname}
    %
    %
    %
	We performed extensive experimental analysis to understand the proposed \modelname~algorithm, including (i) the contributions of different components, (ii) its ability to overcome sparse rewards, (iii) hyperparameter analysis, (iv) its strengths and weaknesses compared to traditional KBC methods, and (v) its running time. 
	First, we used ablation studies to analyze the contributions of different components in \modelname.	To understand the contribution of the proposed neural architecture in \modelname, we created a method, PG-Walk, which uses the same neural architecture as \modelname~but with the same training (PG) and testing (beam search) algorithms as MINERVA \cite{GoforaWalk}. We observed that the novel neural architecture of \modelname~contributes an overall $1\%$ gain relative to MINERVA on NELL995, and it is still $1\%$ worse than \modelname, which uses MCTS for training and testing. To further understand the contribution of MCTS, we created another method, Q-Walk, which uses the same model architecture as \modelname~except that it is trained by Q-learning only without MCTS. Note that this lost about $2\%$ in overall performance on NELL995. We observed similar trends on WN18RR. In addition, we also analyze the importance of MCTS in the testing stage in Appendix \ref{Appendix:MCTSTesting}.

\begin{figure}[t!]
		\centering
		\subfigure[\scriptsize \#Train Rollouts = 16]{
			\includegraphics[width=0.22\textwidth]{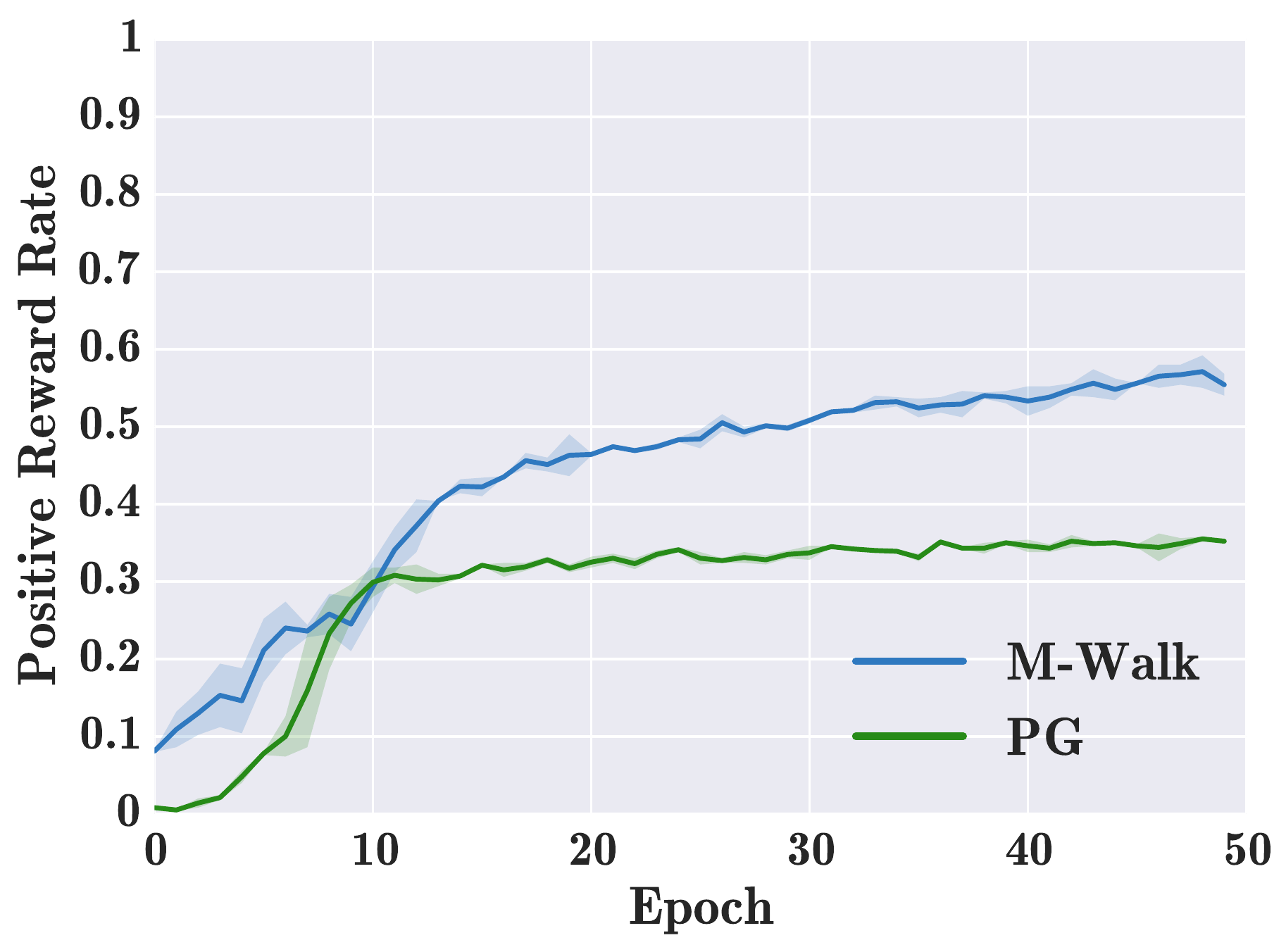}
		}
		\subfigure[\scriptsize \#Train Rollouts = 32]{
			\includegraphics[width=0.22\textwidth]{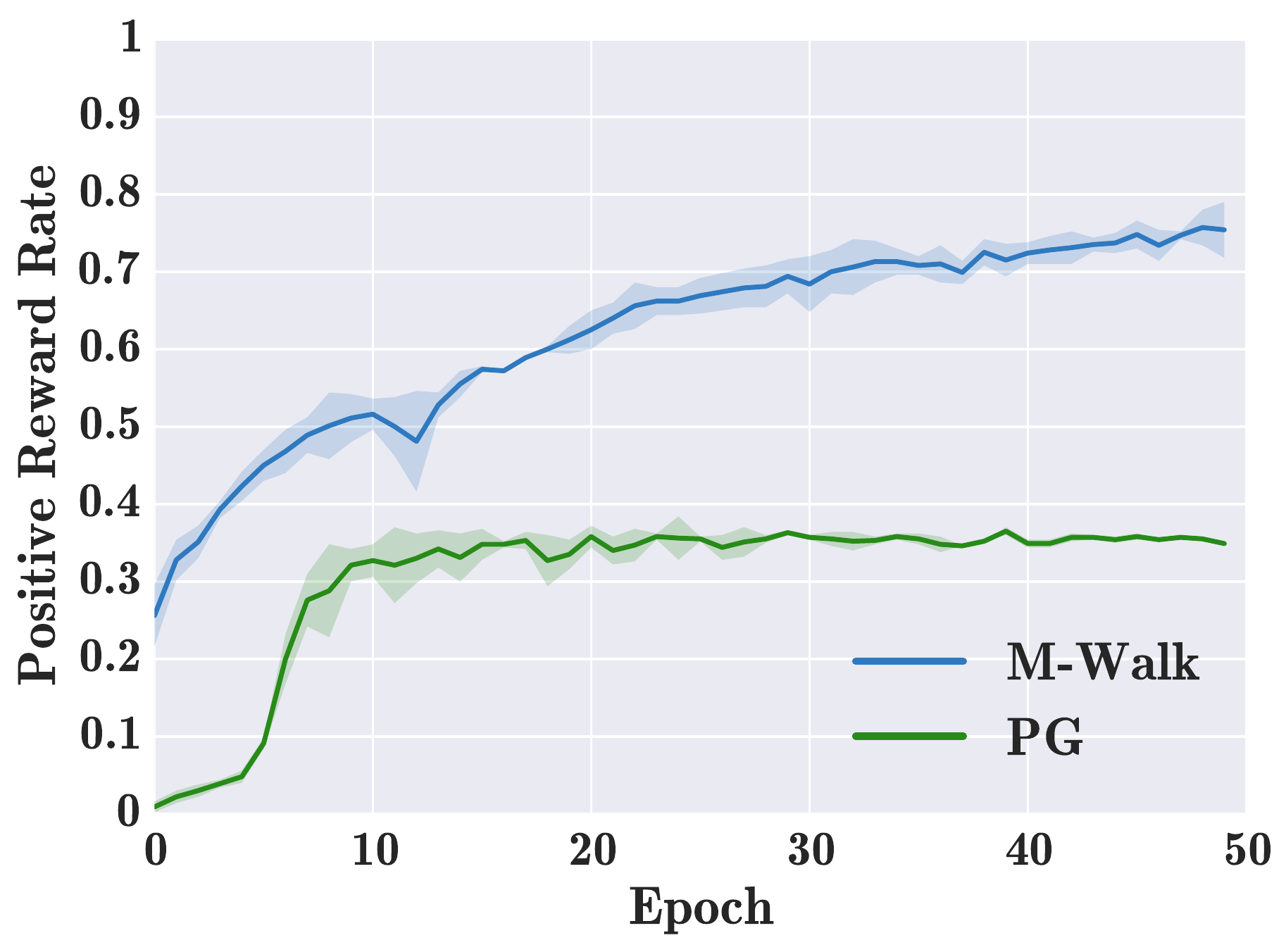}
		}
		\subfigure[\scriptsize \#Train Rollouts = 64]{
			\includegraphics[width=0.22\textwidth]{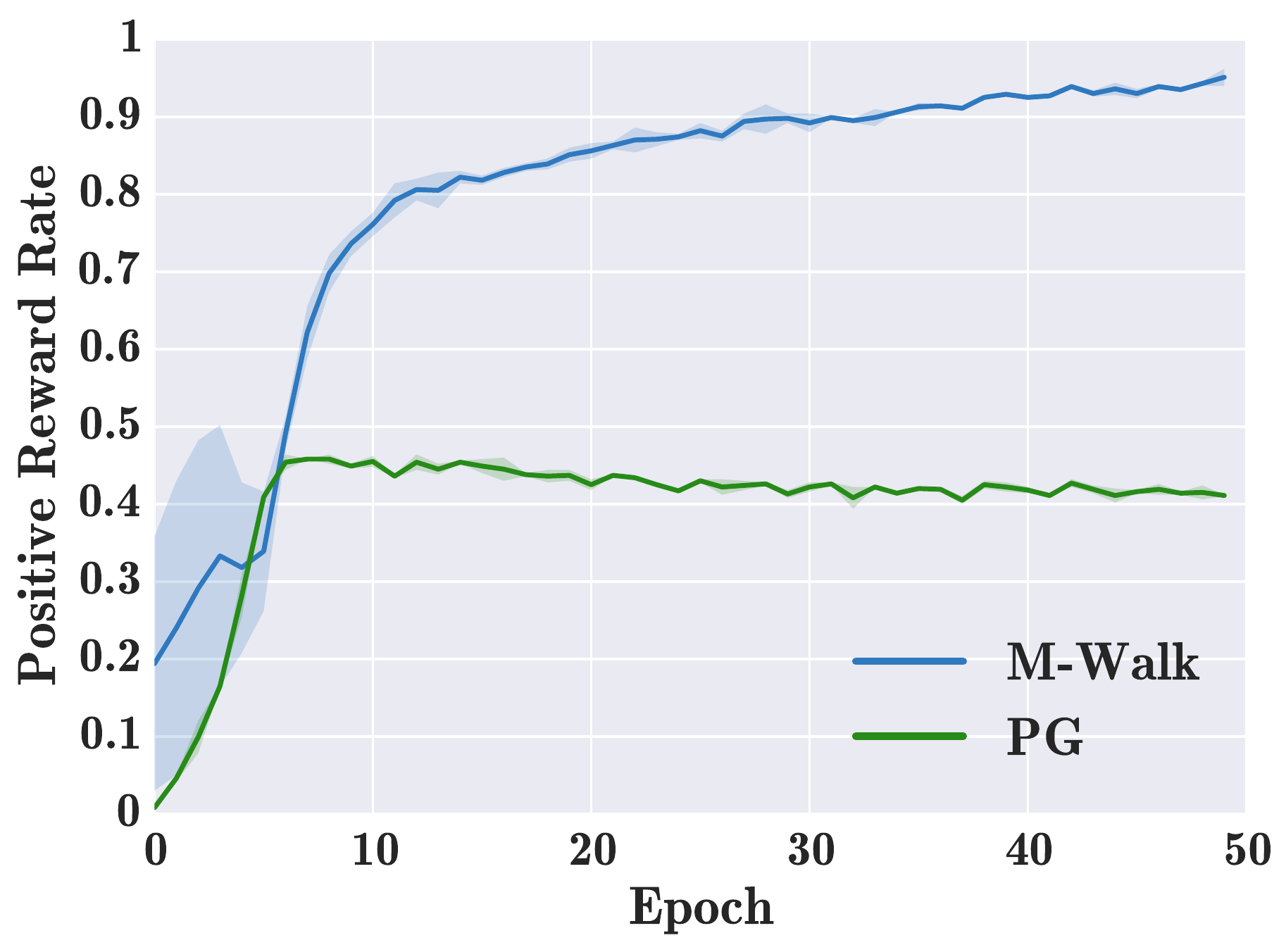}
		}
		\subfigure[\scriptsize \modelname~MCTS Comparison]{
			\includegraphics[width=0.22\textwidth]{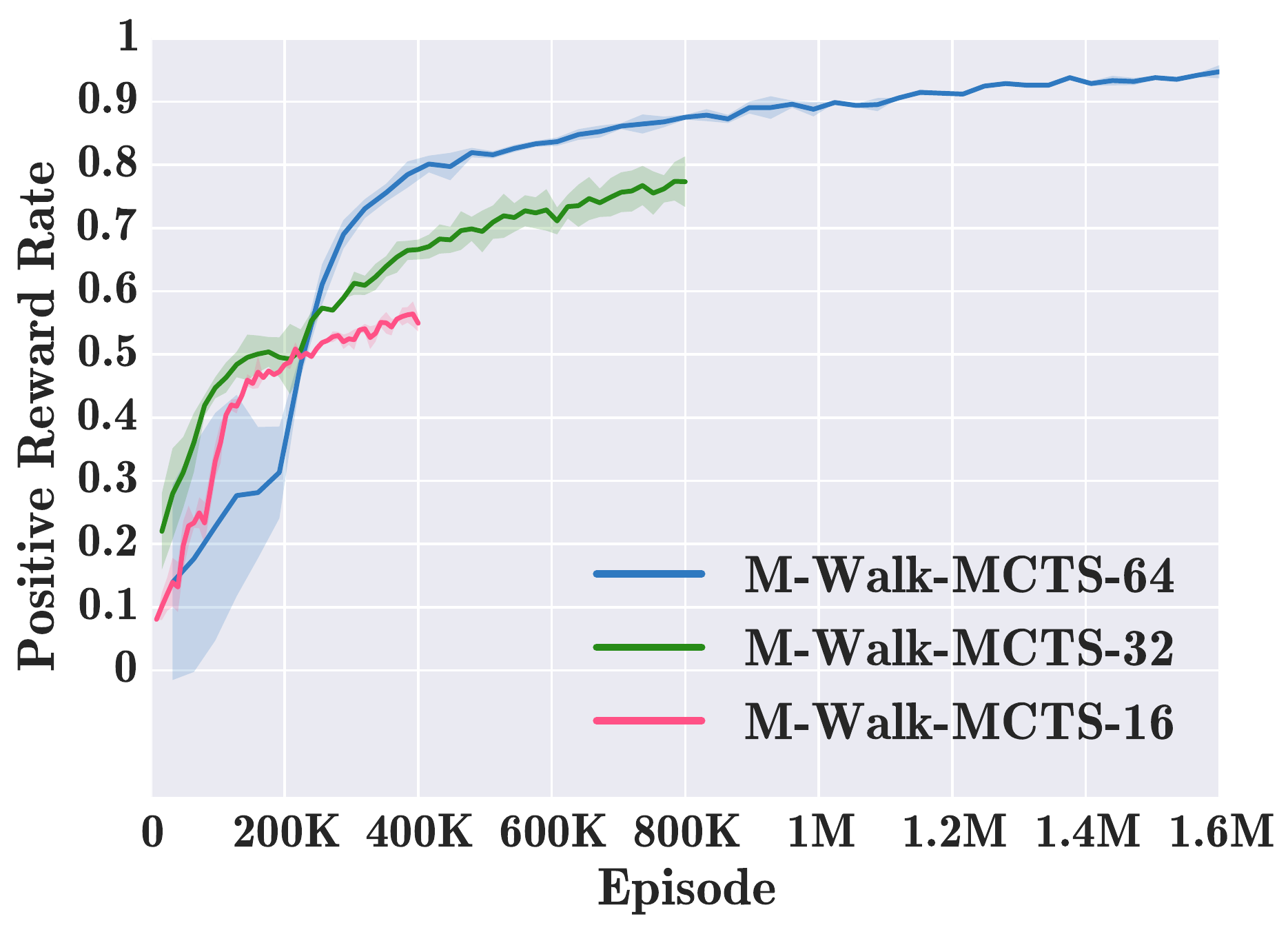}
		}
		\centering
		\subfigure[\scriptsize OrganizationHiredPerson]{
			\includegraphics[width=0.22\textwidth]{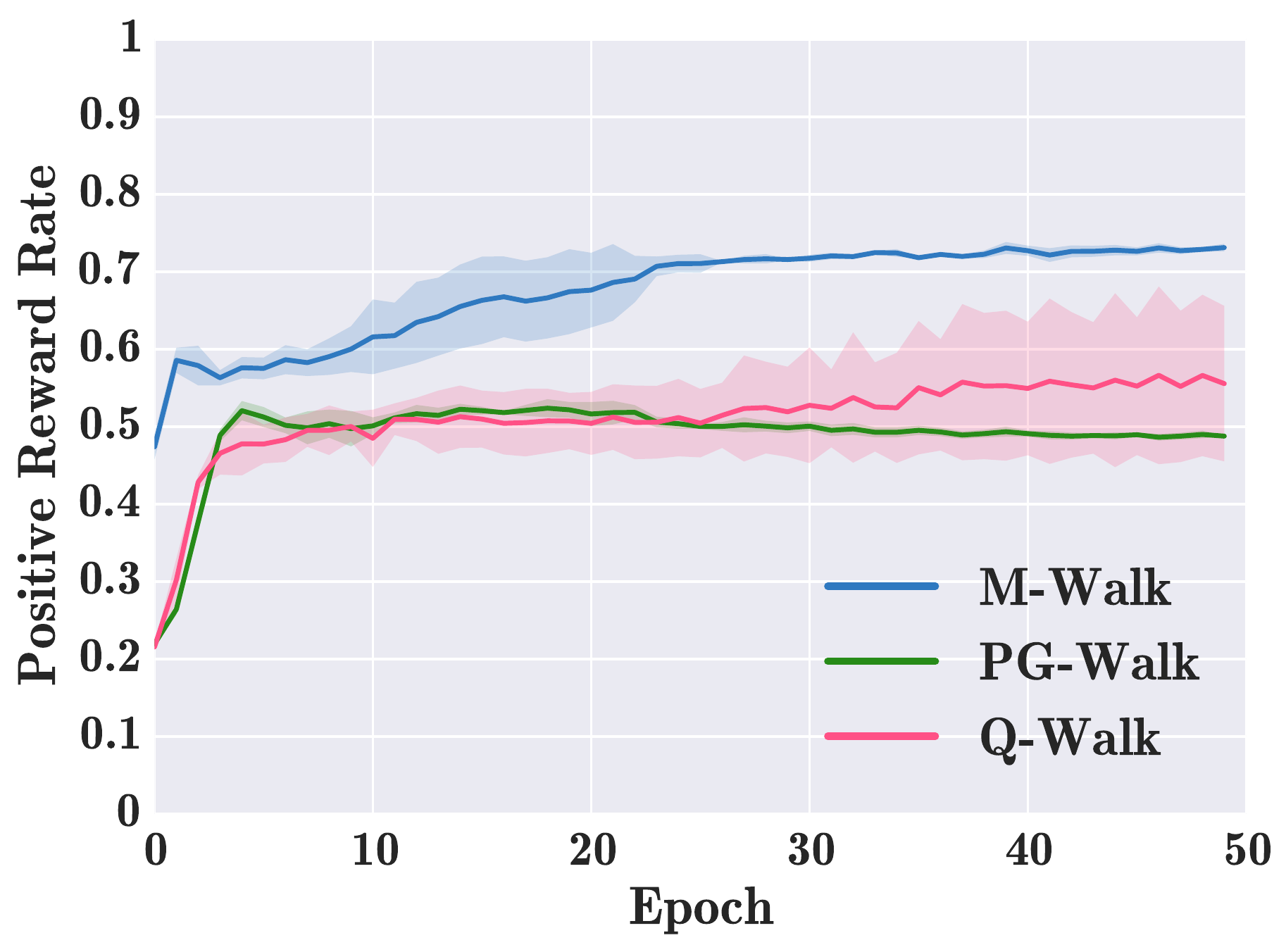}
		}
		\subfigure[\scriptsize AthletePlaysForTeam]{
			\includegraphics[width=0.22\textwidth]{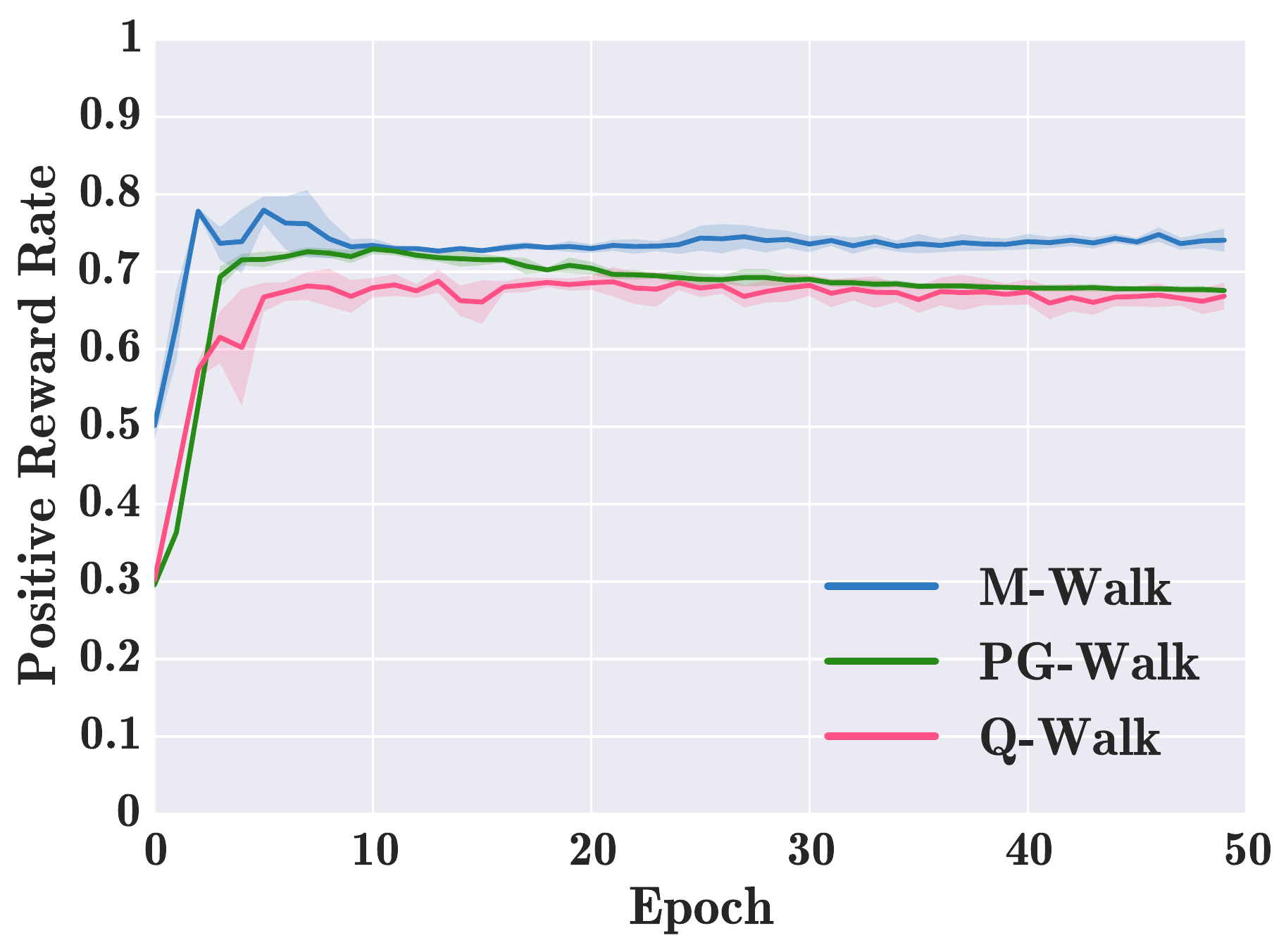}
		}
		\subfigure[\scriptsize PersonLeadsOrganization]{
			\includegraphics[width=0.22\textwidth]{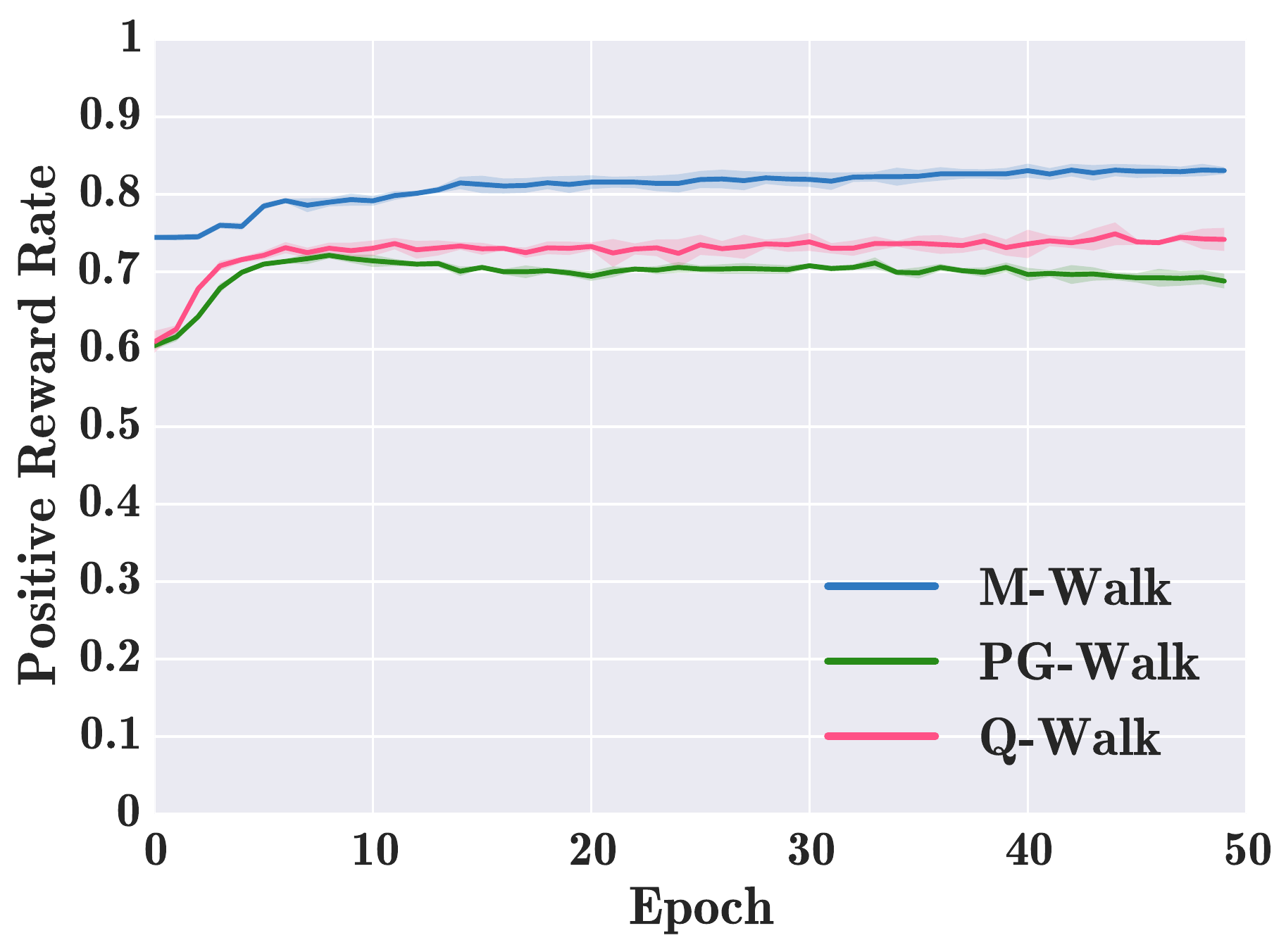}
		}
		\subfigure[\scriptsize WorksFor]{
			\includegraphics[width=0.22\textwidth]{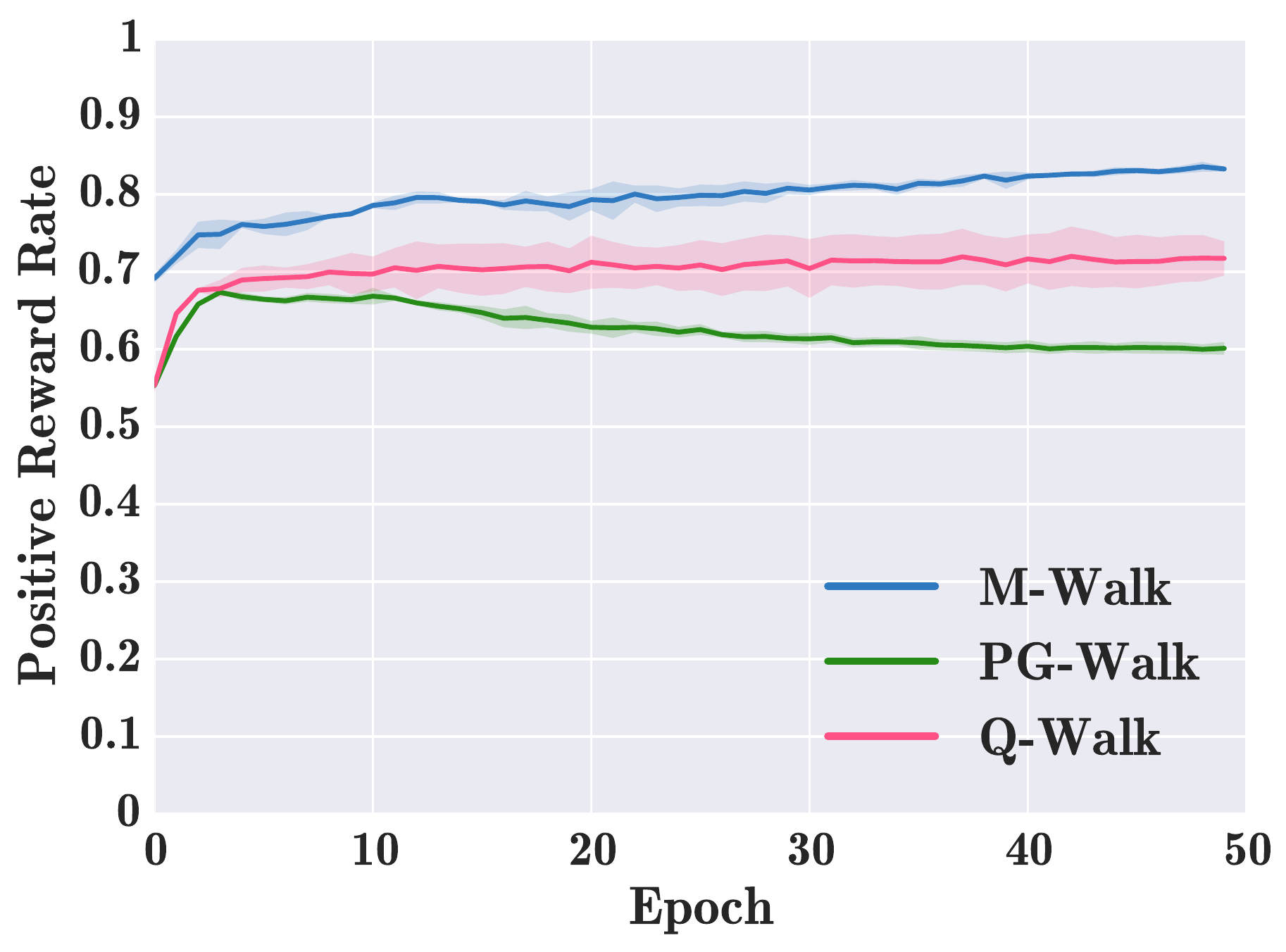}
		}
		\caption{{\small The positive reward rate. Figures (a)-(d) are the results on the Three Glass Puzzle task and Figures (e)-(h) are the results on the NELL-995 task. (See Appendix \ref{Appendix:kbc} for more results.)}}
		\label{fig:kbc_train_success_train}
	\end{figure}

	Second, we analyze the ability of \modelname~to overcome the sparse-reward problem. In Figure \ref{fig:kbc_train_success_train}, we show the positive reward rate (i.e., the percentage of trajectories with positive reward during training) on the Three Glass Puzzle task and the NELL995 tasks. Compared to the policy gradient method (PG-Walk), and Q-learning method (Q-Walk) methods under the same model architecture, \modelname~with MCTS is able to generate trajectories with more positive rewards, and this continues to improve as training progresses. This confirms our motivation of using MCTS to generate higher-quality trajectories to alleviate the sparse-reward problem in graph walking. 
	
\begin{figure}[t!]
	\centering
	\subfigure[\scriptsize MCTS hyperparameter analysis]{
		\label{fig:hyperparameter_hit1_analysis}
		\includegraphics[width=0.31\textwidth]{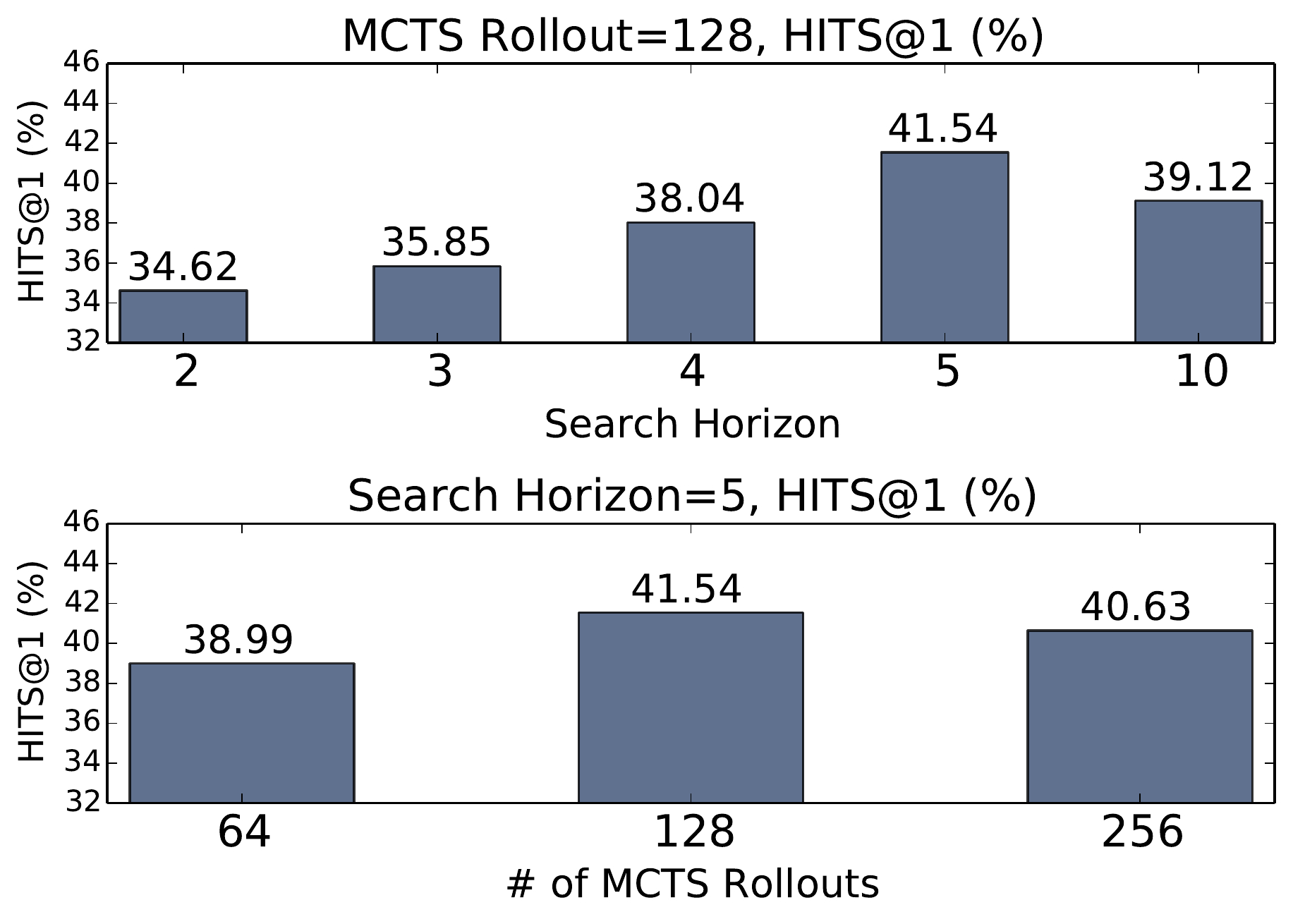}
	}
	\subfigure[\scriptsize HITS@1 (\%) for different path lengths]{
		\label{fig:hit1_bfs}
		\includegraphics[width=0.29\textwidth]{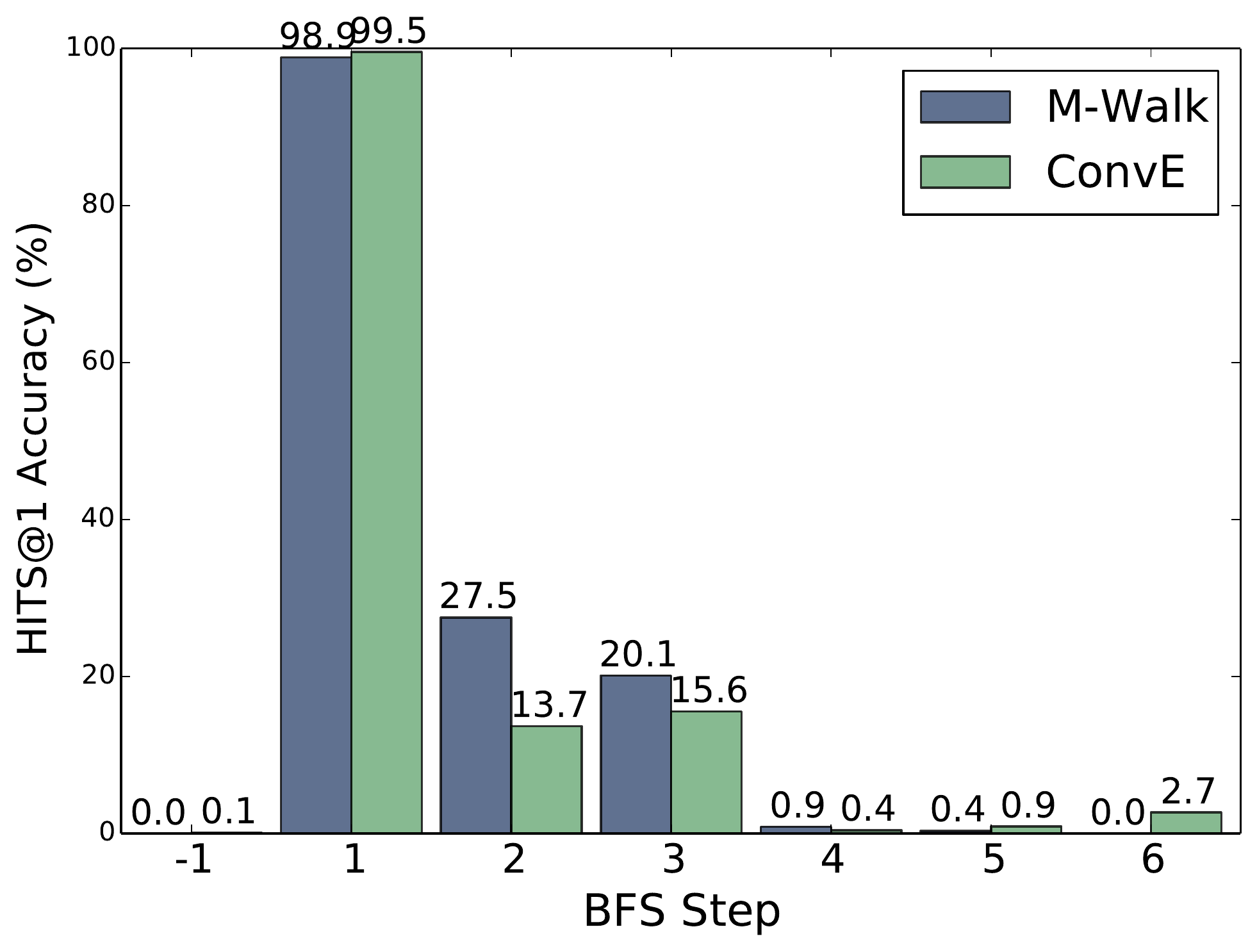}
	}
	\subfigure[\scriptsize Error pattern analysis]{
		\label{fig:error_analysis}
		\includegraphics[width=0.31\textwidth]{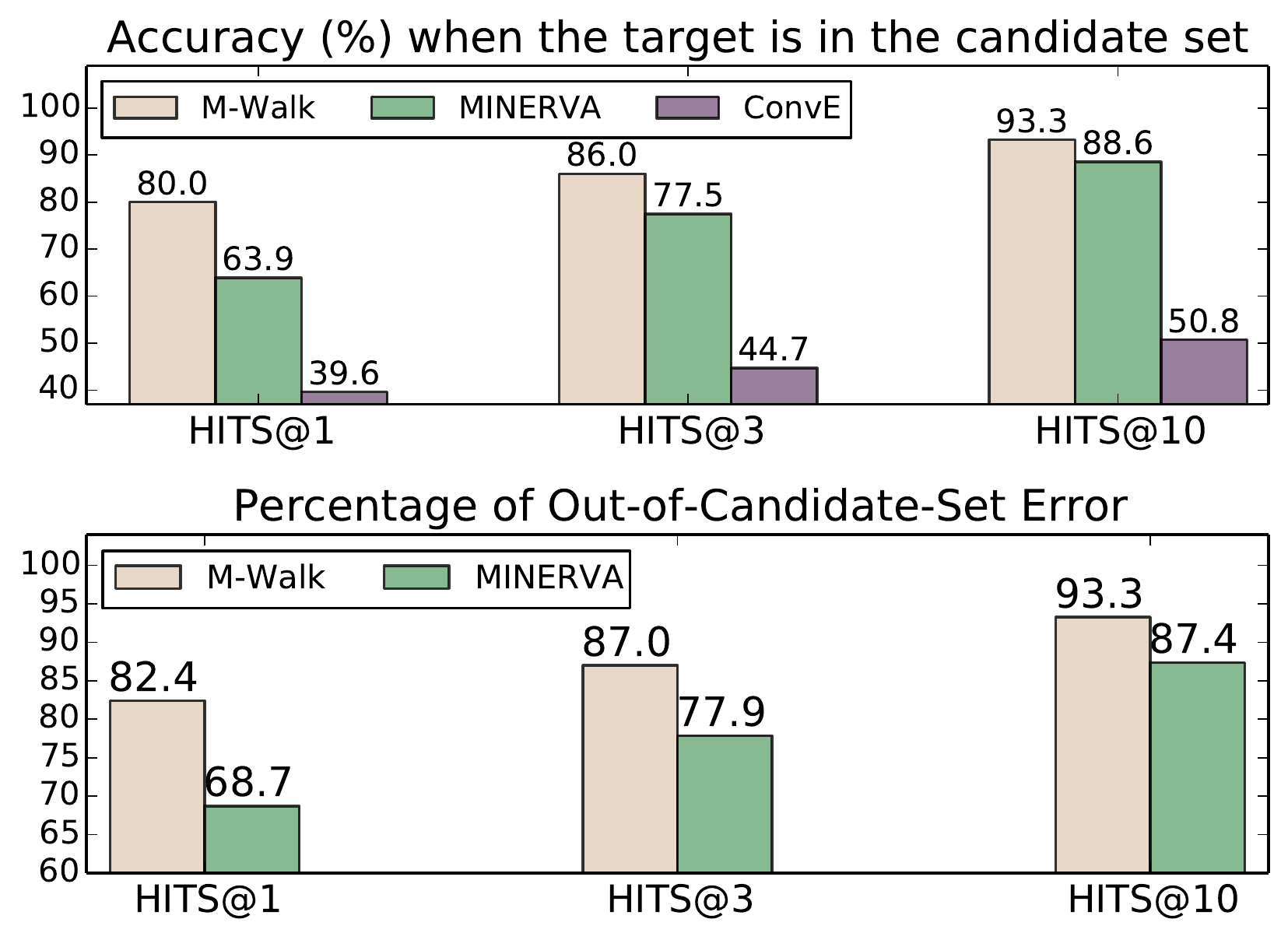}
	}
	\caption{{\small \modelname~hyperparameter and error analysis on WN18RR.
	}}
	\label{fig:wn18rr_analysis}
	\end{figure}

	Third, we analyze the performance of \modelname~under different numbers of MCTS rollout simulations and different search horizons on WN18RR dataset, with results shown in Figure \ref{fig:hyperparameter_hit1_analysis}. We observe that the model is less sensitive to search horizon and more sensitive to the number of MCTS rollouts. 
	Finally, we analyze the strengths and weaknesses of \modelname~relative to traditional methods on the WN18RR dataset. The first question is how \modelname~performs on reasoning paths of different lengths compared to baselines. To answer this, we analyze the HITS@1 accuracy against ConvE in Fig. \ref{fig:hit1_bfs}. We categorize each test example using the BFS (breadth-first search) steps from the query entity to the target entity (-1 means not reachable). We observe that \modelname~outperforms the strong baseline ConvE by 4.6--10.9\% in samples that require 2 or 3 steps, while it is nearly on par for paths of length one. Therefore, \modelname~does better at reasoning over longer paths than ConvE. Another question is what are the major types of errors made by \modelname. Recall that \modelname~only walks through a subset of the graph and ranks a subset of candidate nodes (e.g., MCTS produces about 20--60 unique candidates on WN18RR). When the ground truth is not in the candidate set, \modelname~always makes mistakes and we define this type of error as \emph{out-of-candidate-set error}. To examine this effect, we show in Figure \ref{fig:error_analysis}-top the HITS@K accuracies when the ground truth is in the candidate set.\footnote{The ground truth is always in the candidate list in ConvE, as it examines all the nodes.} It shows that \modelname~has very high accuracy in this case, which is significantly higher than ConvE (80\% vs 39.6\% in HITS@1). We further examine the percentage of out-of-candidate-set errors among all errors in Figure \ref{fig:error_analysis}-bottom. It shows that the major error made by \modelname~is the out-of-candidate-set error. These observations point to an important direction for improving \modelname~in future work: increasing the chance of covering the target by the candidate set.

    
\vspace{-0.5em}
\begin{table}[h]
\begin{center}
\caption{{\small Running time of M-Walk and MINERVA for different combinations of (horizon, rollouts).}}
\label{tab:running_time}
\scriptsize
\begin{tabular}{|c|c|c|c|c|c|}
\hline
Model & M-Walk (5,64) &  M-Walk (5,128) &  M-Walk (3,64) &  M-Walk (3,128) & MINERVA (3,100), best \\
\hline
Training (hrs.)  & $8$  & $14$ & $5$ & $8$ & $3$ \\
\hline
Testing (sec/sample) & $3\times 10^{-3}$ & $6\times 10^{-3}$ & $1.6\times 10^{-3}$ & $2.7\times 10^{-3}$ & $2\times10^{-2}$  \\
\hline
\end{tabular}
\end{center}
\label{default}
\end{table}%

\begin{table*}[!t]
		\centering
		\small
		\caption{Examples of reasoning paths found by \modelname~on the NELL-995 dataset for the relation ``AthleteHomeStadium''. True (False) means the prediction is correct (wrong).}
		\label{tab:reasoning_path_kbc}
		 \resizebox{\columnwidth}{!}{%
		\begin{tabular}{l l}\hline
		\vspace{1mm}
			\textbf{AthleteHomeStadium:}
			\\ 
			\emph{Example 1}: \textsf{athlete ernie banks} $\xrightarrow{\text{AthleteHomeStadium}}$? \vspace{-1mm}
			\\ \\
            \textsf{athlete ernie banks}  $\xrightarrow{\text{AthletePlaysInLeague}}$
			\textsf{SportsLeague mlb} $\xrightarrow{\text{TeamPlaysInLeague}^{-1}}$ \textsf{SportsTeam chicago cubs} $\xrightarrow{\text{TeamHomeStadium}}$
			\textsf{StadiumOrEventVenue wrigley field}, (True)
            \\ \\			
			\emph{Example 2}: \textsf{coach jim zorn} $\xrightarrow{\text{AthleteHomeStadium}}$? \vspace{-1mm}
			\\ \\
            \textsf{coach jim zorn}  $\xrightarrow{\text{CoachWonTrophy}}$
			\textsf{AwardTrophyTournament super bowl} $\xrightarrow{\text{TeamWonTrophy}^{-1}}$ \textsf{SportsTeam redskins} $\xrightarrow{\text{TeamHomeStadium}}$
			\textsf{StadiumOrEventVenue fedex field}, (True)
            \\ \\			
			\emph{Example 3}: \textsf{athlete oliver perez} $\xrightarrow{\text{AthleteHomeStadium}}$? \vspace{-1mm}
			\\ \\
            \textsf{athlete oliver perez}  $\xrightarrow{\text{AthletePlaysInLeague}}$
			\textsf{SportsLeague mlb} $\xrightarrow{\text{TeamPlaysInLeague}^{-1}}$ \textsf{SportsTeam chicago cubs} $\xrightarrow{\text{TeamHomeStadium}}$
			\textsf{StadiumOrEventVenue wrigley field}, (False)
			\\\hline
		\end{tabular}
		}%
	\end{table*}

In Table \ref{tab:running_time}, we show the running time of M-Walk (in-house C++ \& Cuda) and MINERVA (TensorFlow-gpu) for both training and testing on WN18RR with different values of search horizon and number of rollouts (or MCTS simulation number). Note that the running time of M-Walk is comparable to that of MINERVA. Additional results can be found in Figure \ref{fig:hit1_time} of the supplementary material. Finally, in Table \ref{tab:reasoning_path_kbc}, we show examples of reasoning paths found by \modelname.\footnote{More examples can be found in Appendix \ref{Appendix:three_glass_puzzle_paths} of the supplementary material.}

	\section{Related Work}
	\label{Sec:RelatedWork}
	
	\paragraph{Reinforcement Learning}
	Recently, deep reinforcement learning has achieved great success in many artificial intelligence problems \cite{mnih2015human,silver2016mastering, silver2017mastering}. The use of deep neural networks with RL allows policies to be learned from raw data (e.g., images) in an end-to-end manner. Our work also aligns with this direction. Furthermore, the idea of using an RNN to encode the history of observations also appeared in \cite{HausknechtS15, wierstra2010recurrent}. The combination of  model-based and model-free information in our work shares the same spirit as \cite{silver2016mastering, silver2017mastering, pmlr-v70-silver17a, Imagination-Augmented}. Among them, the most relevant are \cite{silver2016mastering, silver2017mastering}, which combine MCTS with neural policy and value functions to achieve superhuman performance on Go. Different from our work, the policy and the value networks in \cite{silver2016mastering} are trained separately without the help of MCTS, and are only used to help MCTS after being trained. The work \cite{silver2017mastering} uses a new policy iteration method that combines the neural policy and value functions with MCTS during training. However, the method in \cite{silver2017mastering} improves the policy network from the MCTS probabilities of the moves, while our method improves the policy from the trajectories generated by MCTS. Note that the former is constructed from the visit counts of all the edges connected to the MCTS root node; it only uses information near the root node to improve the policy. By contrast, we improve the policy by learning from the trajectories generated by MCTS, using information over the entire MCTS search tree.
	
	\paragraph{Knowledge Base Completion}
	In KBC tasks, early work \cite{bordes2013translating} focused on learning vector representations of entities and relations. 
	Recent approaches have demonstrated limitations of these prior approaches: they suffer from cascading errors when dealing with compositional (multi-step) relationships \cite{GuuMiLi15}.
	Hence, recent works \cite{gardner2014incorporating, neelakantan2015compositional, GuuMiLi15, PTransE2015EMNLP, KristinaQu16} have proposed approaches for injecting multi-step paths such as random walks through sequences of triples during training, further improving performance on KBC tasks.
	IRN \cite{shen2017modeling} and Neural LP \cite{yang2017differentiable} explore multi-step relations by using an RNN controller with attention over an external memory. 
	Compared to RL-based approaches, it is hard to interpret the traversal paths, and these models can be computationally expensive to access the entire graph in memory \cite{shen2017modeling}.
	Two recent works, DeepPath \cite{DeepPath} and MINERVA \cite{GoforaWalk}, use RL-based approaches to explore paths in knowledge graphs. 
	DeepPath requires target entity information to be in the state of the RL agent, and cannot be applied to tasks where the target entity is unknown.
	MINERVA \cite{GoforaWalk} uses a policy gradient method to explore paths during training and test.
	Our proposed model further exploits state transition information by integrating the MCTS algorithm. 
	Empirically, our proposed algorithm outperforms both DeepPath and MINERVA in the KBC benchmarks.\footnote{A preliminary version of M-Walk with limited experiments was reported in the workshop paper \cite{shen2018reinforcewalk}.}


	\section{Conclusion and Discussion}
	\label{Sec:Conclusion}
	
	We developed an RL-agent (\modelname) that learns to walk over a graph towards a desired target node for given input query and source nodes. Specifically, we proposed a novel neural architecture that encodes the state into a vector representation, and maps it to Q-values and a policy. To learn from sparse rewards, we propose a new reinforcement learning algorithm, which alternates between an MCTS trajectory-generation step and a policy-improvement step, to iteratively refine the policy. At test time, the learned networks are combined with MCTS to search for the target node. Experimental results on several benchmarks demonstrate that our method learns better policies than other baseline methods, including RL-based and traditional methods on KBC tasks. Furthermore, we also performed extensive experimental analysis to understand \modelname. We found that our method is more accurate when the ground truth is in the candidate set. We also found that the out-of-candidate-set error is the main type of error made by M-Walk. Therefore, in future work, we intend to improve this method by reducing such out-of-candidate-set errors.
	
	\subsection*{Acknowledgments}
    We thank Ricky Loynd, Adith Swaminathan, and anonymous reviewers for their valuable feedback.

	\bibliography{ICML2018_Ref}
	\bibliographystyle{plain}

	\clearpage
	
	\appendix

\section{Derivation of the recursion for $q_t$}
\label{Appendix:qt_derivation}

Recalling the definition $q_t \defeq s_{t-1} \cup \{a_{t-1},n_t\}$ and using the recursion \eqref{Equ:\modelname:s_t1}, we have
    \begin{align}
        q_{t+1}     &\overset{(a)}{=}  
                            s_t \cup \{a_t, n_{t+1} \}
                            \nn\\
                    &\overset{(b)}{=}
                            s_{t-1} \cup \{a_{t-1}, n_t, \mc{E}_{n_t}, \mc{N}_{n_t} \}
                            \cup \{ a_t, n_{t+1} \}
                            \nn\\
                    &\overset{(c)}{=}
                            q_t \cup \{ \mc{E}_{n_t}, \mc{N}_{n_t}, a_t, n_{t+1} \}
                            \nn
    \end{align}
where step (a) uses the definition of $q_{t+1}$, step (b) substitutes the recursion \eqref{Equ:\modelname:s_t1}, and step (c) uses the definition of $q_t$.

\section{Algorithm Implementation Details}
\label{Appendix:AlgDetails}

The detailed algorithm of \modelname~is described in Algorithm \ref{alg:reinforcewalk}. 
\begin{algorithm}[h]
	\begin{algorithmic}[1]
	\STATE {\bfseries Input: }{Graph $\mc{G}$; Initial node $n_S$; Query $q$;  Target node $n_{T}$; Maximum Path Length $T_{\text{max}}$; MCTS Search Number $E$;}
	\FOR{episode $e$ in $[1..E]$}
		\STATE Set current node $n_0 = n_S$; $q_{0} = f_{\theta_q}(q, 0, 0, n_0)$\;
		\FOR{$t=0\ldots T_{\text{max}}$}

		\STATE Lookup from dictionary to obtain $W(s_t, a)$ and $N(s_t, a)$ \;
		\STATE Select the action $a_{t}$ with the maximum PUCT value:
		\begin{align}
		a_t =
		\argmax_a \! \left\{ \! c \!\cdot\! \pi_{\theta}(a | s_t)^{\beta} \frac{\sqrt{\sum_{a'} N(s_t, a')}}{1 \!+\! N(s_t, a) } \!+\! \frac{W(s_t, a)} {N(s_t, a)}
		\!\right\} \nonumber
		\end{align}
		\STATE Update $q_{t+1} = f_{\theta_q}(q_t, h_{A, t}, h_{a_t, t}, n_{t+1})$\;
		\IF{$a_t$ is STOP}	

		\STATE Compute estimated reward value $V_{\theta}(s_t) = Q(s_t, a_t=\tt{STOP})$ \;
		\STATE Add generated path $p$ into a path list \;
		\STATE Backup along the path $p$ to update the visit count $N(s_t, a)$ using \eqref{Equ:Appendix:MCTS_backup_N} and the total action reward $W(s_t, a)$ using \eqref{Equ:Appendix:MCTS_backup_W} on the $(s_t, a)$-th edge on the MCTS tree \;
		\STATE {\bf Break}
		\ENDIF
		\ENDFOR
		\ENDFOR
	\FOR{each path $p$ in the path list}
		\STATE Set reward $r = 1$ if the end of the path $n_t = n_{T}$ otherwise $r=0$\;
		\STATE Repeatedly update the model parameters with Q-learning:
		\begin{align}
		\theta	&\leftarrow	\theta  + \alpha \cdot \nabla_{\theta} Q_{\theta}(s_t,a_t) \times
		\left( r(s_t,a_t)
		+ \gamma \max_{a'} Q_{\theta}(s_{t+1},a') - Q_{\theta}(s_t,a_t) \right) \nonumber
		\end{align}
	\ENDFOR	
	\caption{\modelname~Training Algorithm}
	\label{alg:reinforcewalk}
	\end{algorithmic}
\end{algorithm}

\subsection{MCTS implementation}
\label{Appendix:AlgDetails:MCTS}
In the MCTS implementation, we maintain a lookup table to record values $W(s_t, a)$ and $N(s_t, a)$ for each visited state-action pair. The state $s_t$ in the graph walk problem contains all the information along the traversal path, and $n_t$ is the node at the current step $t$.
We assign an index $i_a$ to each candidate action $a$ from $n_t$, indicating that $a$ is the $i_a$-th action of the node $n_t$. 
Thus, the state $s_t$ can be encoded as a path string $P_{s_t} = (q, n_0, i_{a_0}, n_1, i_{a_1}, \ldots, n_t)$.
We build a dictionary $\mc{D}$ using the path string as a key, and we record $W(s_t, a) $ and $N(s_t, a)$ as values in $\mc{D}$. 
In the backup stage, the $W(s_t,a)$ and $N(s_t,a)$ values are updated for each state-action pair along with the traversal path in MCTS:
\begin{align}
	N(s_t, a) &= N(s_t, a) + \gamma^{T-t}
	\label{Equ:Appendix:MCTS_backup_N}
	\\
	W(s_t, a) &= W(s_t, a) + \gamma^{T-t} V_{\theta}(s_{T}),
	\label{Equ:Appendix:MCTS_backup_W}
\end{align}
where $T$ is the length of the traversal path, $\gamma$ is the discount factor of the MDP, and $V_{\theta}(s_T)$ is the terminal state-value function modeled by $V_{\theta}(s_T) \defeq Q(s_T, a=\tt{STOP})$.

In our experiments, the softmax temperature parameter $\tau$ in the policy network $\pi_{\theta}$ (see \eqref{Equ:\modelname:PolicyNet}) is set to be a constant. An alternative choice is to anneal it during training (e.g., $\tau = 1 \rightarrow 0$). However, we did not observe this to produce any significant difference in performance in our experiments. We believe the main reason is that $\pi_{\theta}$ is only used as a prior to bias the MCTS search, while the exploration of MCTS is controlled by the parameters $c$ and $\beta$ of \eqref{Equ:\modelname:PUCT}.


\subsection{Experiment details}
\label{Appendix:exp_details}

\subsubsection{Three Glass Puzzle}

	\begin{figure}[h!]
		\centering
		\includegraphics[width=0.5\textwidth]{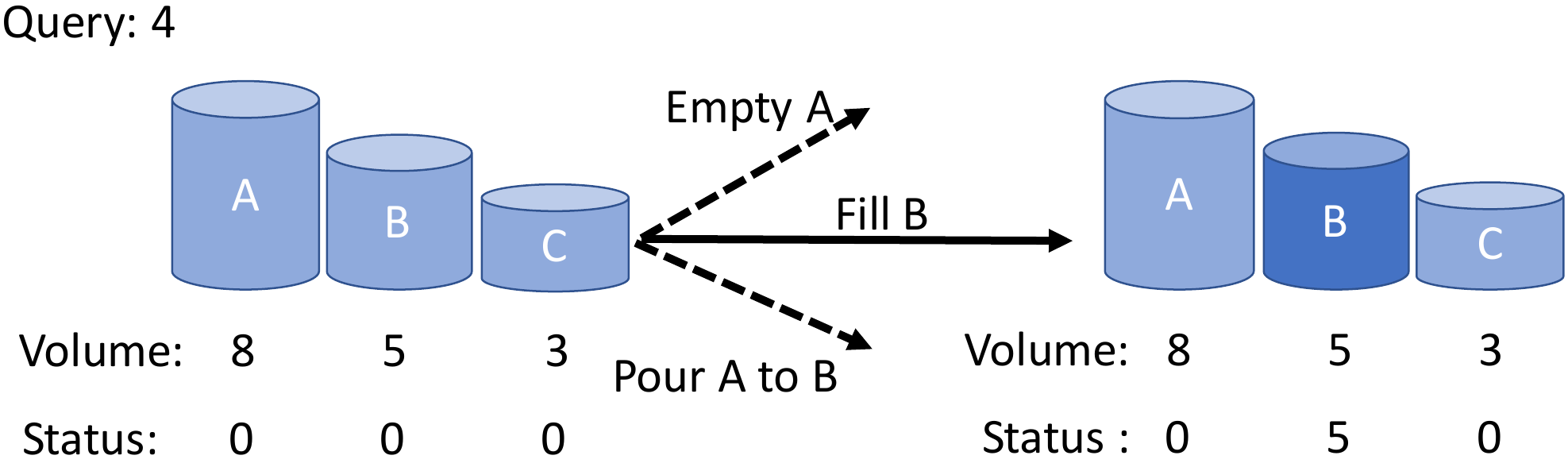}
		\caption{Graph traversal in the Three Glass Puzzle problem.}
		\label{fig:reorder_a}
	\end{figure}

\paragraph{An example} Figure \ref{fig:reorder_a} illustrates one step in solving a Three Glass Puzzle. 
	The following action sequences provide one solution to achieve the target $q=4$, given initially empty containers with capacities  $(A=8, B=5, C=3)$, where $a, b, c$ denote the current contents of the containers:
	\begin{itemize}
		\item  {Initial state} $\rightarrow (a=0, b=0, c=0)$  
		\item  {Fill $\mathcal{B}$} $\rightarrow (a=0, b=5, c=0)$ 
		\item  {Pour from $\mathcal{B}$ to $\mathcal{C}$} $\rightarrow (a=0, b=2, c=3)$ 
		\item  {Empty $\mathcal{C}$} $\rightarrow (a=0, b=2, c=0)$ 
    	\item  {Pour from $\mathcal{B}$ to $\mathcal{C}$} $\rightarrow (a=0, b=0, c=2)$ 
		\item  {Fill $\mathcal{B}$} $\rightarrow (a=0, b=5, c=2)$ 
		\item  {Pour from $\mathcal{B}$ to $\mathcal{C}$} $\rightarrow (a=0, b=4, c=3)$
	\end{itemize}

\paragraph{Data generation} In the Three Glass Puzzle experiments, we randomly draw four integers from $[1, 50)$ to represent the capacities $A, B, C$, and the desired volume $q$. 
We further restrict the values so that $A \geq B \geq C$ and $q < A$, to avoid data duplication. We discard puzzles for which there is no solution. Finally, we keep 600 unique puzzles as the experimental dataset, where 500 puzzles are used for training and the other 100 are used to test a model's generalization capability on the unseen test set.

\paragraph{Experiment settings and hyperparameters}
Let $a, b, c$ be the current status of each container, and define the puzzle status at step $t$ as $n_t = [I_A^T, I_B^T, I_C^T, I_a^T, I_b^T, I_c^T]^T$, where $I_{x}$ is the one-hot representation to encode the value of $x$. 
Given that $A, B, C, a, b$ and $c$ are all smaller than $50$ in the experiment, the dimension of $n_t$ is 300. 
The initial query $q$ is obtained by $q = {E_{mb}}[q]$, where ${E_{mb}}$ is a query embedding lookup table and ${E_{mb}}[x]$ indicates the $x$-th column.
The query embedding dimension is set to $64$.
In the Three Glass Puzzle, there are 13 actions in total: fill one container to its capacity, empty one container, pour one container into another container, and a STOP action to terminate the game. 
We set the maximum length of an action sequence (i.e., the search horizon) to be $12$, where only the STOP action can be taken on the final step.
After the STOP action has been taken, the system evaluates the action sequence and assigns a reward $r = 1$ if the final status is a success, otherwise $r = 0$. The $f_{{\theta}_{S}}$ and $f_{{\theta}_{A}}$ functions are modeled by two different DNNs with the same architecture: two fully-connected layers with 32 hidden dimensions and ReLU activation function.
$f_{{\theta}_{v}}$ is two fully-connected layers with 16 hidden dimensions, where the first hidden layer uses a ReLU activation function and the output layer uses a linear activation function. 
$f_{{\theta}_{q}}$ is modeled by a GRU with hidden size 64.
The hyperparameters in PUCT are set to $c=0.5$ and $\beta=0.2$. We use the ADAM optimization algorithm with learning rate $0.0005$ during training, and we set the mini-batch size to $8$.

\begin{table}[t]
	\centering
	{
		\caption{A List of actions for each container in the Three Glass Puzzle. The agent can also determine to take the STOP action to terminate the game.}
		\label{tab:puzzle_actions}
		\begin{tabular}{|c|c|c|c|}
			\hline
			 Empty $\mathcal{A}$ & Fill $\mathcal{A}$ & Pour $\mathcal{A}$ to $\mathcal{B}$  & Pour $\mathcal{A}$ to $\mathcal{C}$ \\
			 \hline
			 Empty $\mathcal{B}$ & Fill $\mathcal{B}$ & Pour $\mathcal{B}$ to $\mathcal{A}$  & Pour $\mathcal{B}$ to $\mathcal{C}$ \\
			 \hline
			 Empty $\mathcal{C}$ & Fill $\mathcal{C}$ & Pour $\mathcal{C}$ to $\mathcal{A}$  & Pour $\mathcal{C}$ to $\mathcal{B}$ \\
			\hline
		\end{tabular}
	}
\end{table}

\label{Appendix:puzzle_details}

\subsubsection{Knowledge Base Completion}
\label{Appendix:kbc_details}

\begin{table}[t]
\caption{Knowledge base completion datasets statistics.}
\label{tab:kbc_stats}
\centering
\begin{tabular}{c cccccc}
\toprule
Dataset  & \# Train & \# Test & \# Relation & \# Entity & avg. degree & median degree\\
\midrule
WN18RR & 86,835 & 3,134 & 11 & 40,943 &  2.19 & 2 \\
NELL-995 & 154,213 & 3,992 & 200 & 75,492 & 4.07 & 1 \\
FB15K-237 & 272,115 & 20,466 & 237 & 14,541 & 19.74 & 14\\
\bottomrule
\end{tabular}
\end{table}

\paragraph{Statistics of the three datasets}
The NELL-995 knowledge dataset contains $75,492$ unique entities and $200$ relations. 
WN18RR contains $93,003$ triples with $40,943$ entities and $11$ relations. And FB15k-237, a subset of FB15k where inverse relations are removed, contains $14,541$ entities and $237$ relations. The detailed statstics are shown in Table \ref{tab:kbc_stats}.

\paragraph{Experiment settings and hyperparameters}
For the proposed \modelname, we set the entity embedding dimension to $4$ and relation embedding dimension to $64$.
The maximum length of the graph walking path (i.e., the search horizon) is $8$ in the NELL-995 dataset and $5$ in the WN18RR dataset.
After the STOP action has been taken, the system evaluates the action sequence and assigns a reward $r = 1$ if the agent reaches the target node, otherwise  $r = 0$.
The initial query $q$ is the concatenation of the entity embedding vector and the relation embedding vector. 
The $f_{{\theta}_{S}}$ and $f_{{\theta}_{A}}$ functions are modeled by two different DNNs with the same architecture: two fully-connected layers with 64 hidden dimensions and the ReLU activation function.
$f_{{\theta}_{v}}$ is two fully-connected layers with 16 hidden dimensions, where the first hidden layer uses a Tanh activation function and the output layer uses a linear activation function. 
$f_{{\theta}_{q}}$ is modeled by a GRU with hidden size 64.
The hyperparameters in PUCT are set to $c=2$ and $\beta=0.5$. 
We roll out $32$ MCTS paths in both training and testing in the NELL-995 dataset and $128$ MCTS paths in the WN18RR dataset.
We use the ADAM optimization algorithm for model training with learning rate $0.0001$, and we set the mini-batch size to $8$.


\section{Additional Experiments}
\label{Appendix:AdditionalExperiments}
\subsection{The Three Glass Puzzle task in different settings}
\label{Appendix:MCTSTesting}

	\begin{figure}[t!]
		\centering
		\subfigure[Test Beam / Rollout = 128]{
			\includegraphics[width=0.28\textwidth]{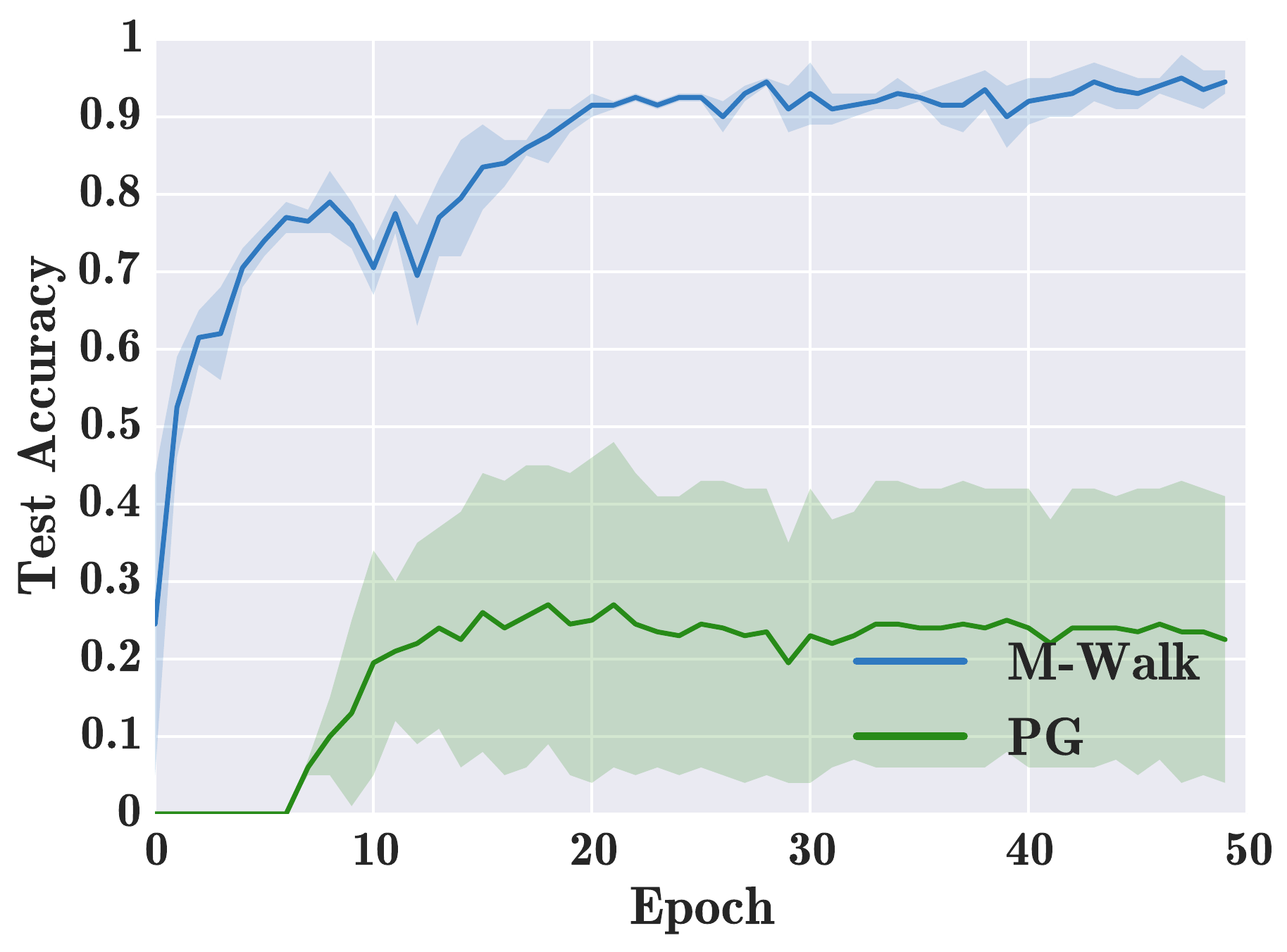}
		}
		\subfigure[Test Beam / Rollout = 300]{
			\includegraphics[width=0.28\textwidth]{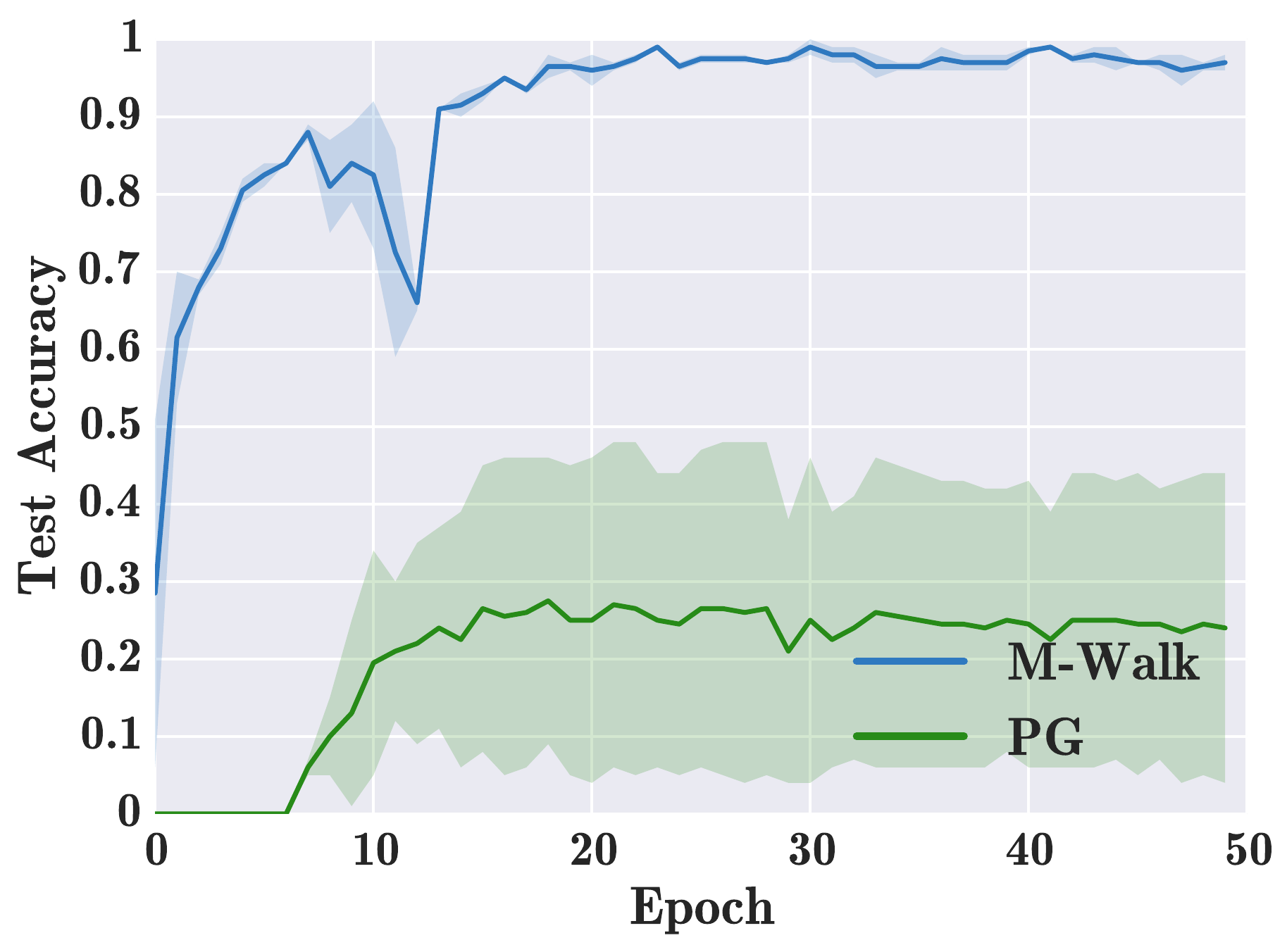}
		}
		\caption{Three Glass Puzzle test accuracy, where ``PG'' stands for policy gradient.}
		\label{fig:puzzle_test_accuracy}
	\end{figure}

	\begin{table}[t]
		\centering
		\caption{Three Glass Puzzle test accuracy (\%), where ``Beam'' denotes beam search.}
		\label{tab:lp_results}
        \resizebox{\columnwidth}{!}{%
		{
\begin{tabular}{|l|c|c|c|c|c|c|c|}
\hline
Size & 1          & 10         & 50         & 100        & 200        & 300        & 400        \\ \hline
\multicolumn{1}{|l|}{PG (Beam)}     & 9.3 (2.1)  & 30.7 (4.5) & 39.3 (3.2) & 45.3 (4.5) & 47.7 (3.2) & 48.7 (3.2) & 49.0 (2.6) \\ \hline
\multicolumn{1}{|l|}{M-Walk (Beam)} & 18.0 (1.7) & 46.0 (7.0) & 60.3 (7.8) & 67.0 (7.0) & 69.0 (6.2) & 69.3 (6.4) & 71.7 (4.5) \\ \hline
\multicolumn{1}{|l|}{M-Walk (MCTS)} & {\bf 18.0} (1.7) & {\bf 63.3} (5.0) & {\bf 84.3} (3.1) & {\bf 90.7} (2.5) & {\bf 95.0} (2.6) & {\bf 96.3} (1.5) & {\bf 99.0} (1.0) \\ \hline
\end{tabular}}
		}
	\end{table}
	
		\begin{table}[t]
		\centering
		{
			\caption{BFS, DFS and \modelname~on Three Glass Puzzle.}
			\label{tab:graph_traversal_results}
			\begin{tabular}{|c|c|c|}
				\hline
				{Method} & {Average \# Steps} & {Max \# Steps} \\
				\hline
				BFS & 264.7 & 1030 \\
				DFS & 192.2 & 1453 \\
				\modelname~& 94.9 & 897 \\
				\hline
			\end{tabular}
		}
	\end{table}
	
We now present more experiments on the Three Glass Puzzle task under different settings. First, to see how fast \modelname~converges, we show in Figure \ref{fig:puzzle_test_accuracy} the learning curves of \modelname~and PG. It shows that \modelname~converges much faster than PG and achieves better results on this task. In Table \ref{tab:lp_results}, we report the test accuracy of \modelname~and vanilla policy gradient (REINFORCE/PG) with different beam search sizes and different MCTS rollouts during testing. The number of MCTS simulations for training \modelname~is fixed to be $32$. 
We observe that \modelname~with MCTS achieves the best test accuracy overall. In addition, with larger beam search sizes and MCTS rollouts, the test accuracy improves substantially. Furthermore, replacing the MCTS in \modelname~by beam search at test time degrades the performance greatly, which shows that MCTS is also very important for \modelname~at test time. 

As mentioned earlier, conventional graph traversal algorithms such as Breadth-First Search (BFS) and Depth-First Search (DFS) cannot be applied to the graph walking problem, because the ground truth target node is not known at test time. However, to understand how quickly \modelname~with MCTS can find the correct target node, we compare it with BFS and DFS in the following cheating setup. Specifically, we apply BFS and DFS to the test set of the Three Glass Puzzle task by disclosing the target node to them. In Table \ref{tab:graph_traversal_results}, we report the average traversal steps and maximum steps to reach the target node. The \modelname~with MCTS algorithm is able to find the target node more efficiently than BFS or DFS.

\subsubsection{Knowledge Graph Link Prediction}
\label{Appendix:kbc}

\begin{figure}[t!]
		\centering
			\subfigure[\scriptsize TeamPlaySports]{
				\includegraphics[width=0.31\textwidth]{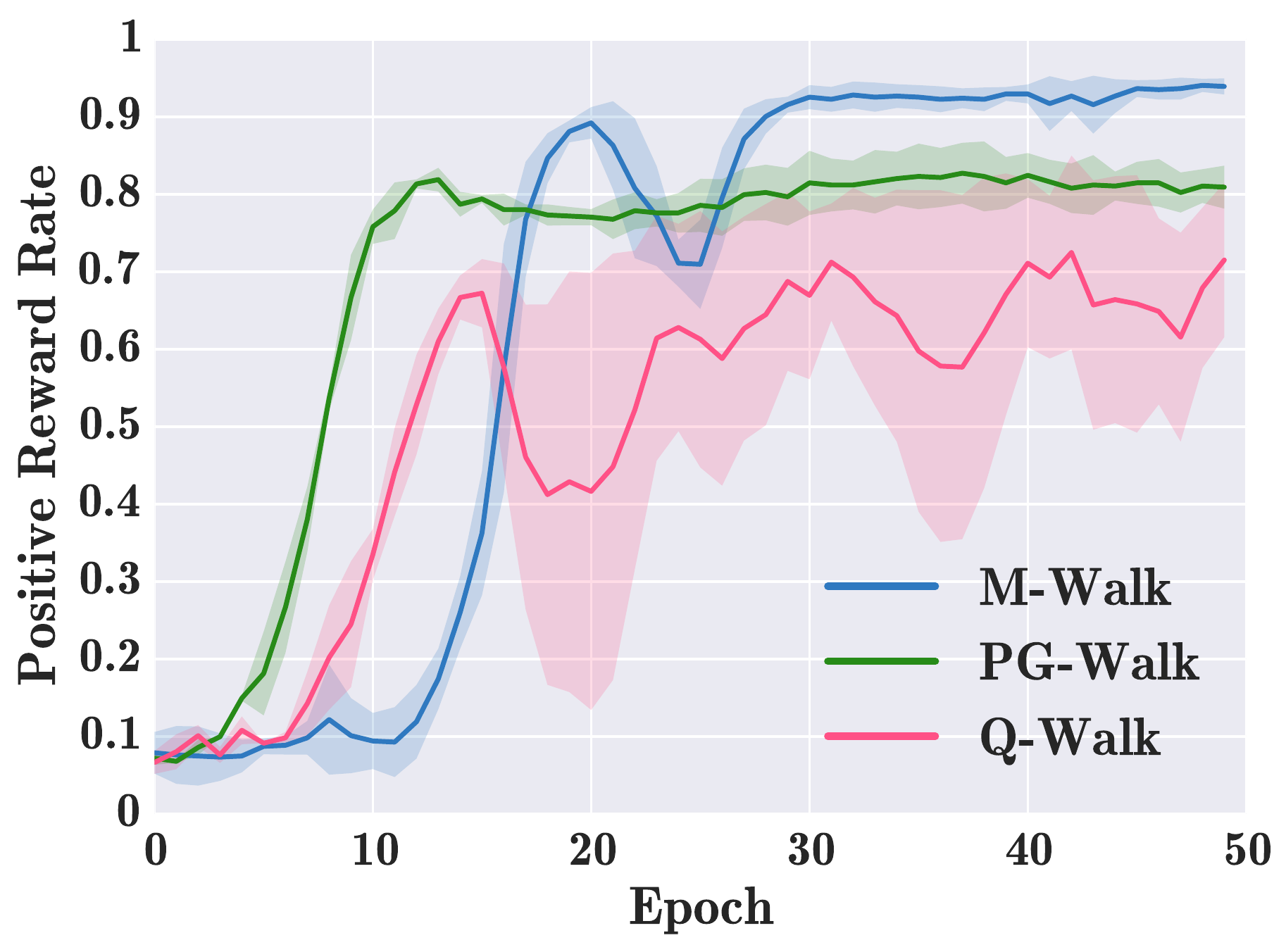}
			}
			\subfigure[\scriptsize AthletePlaysInLeague]{
				\includegraphics[width=0.31\textwidth]{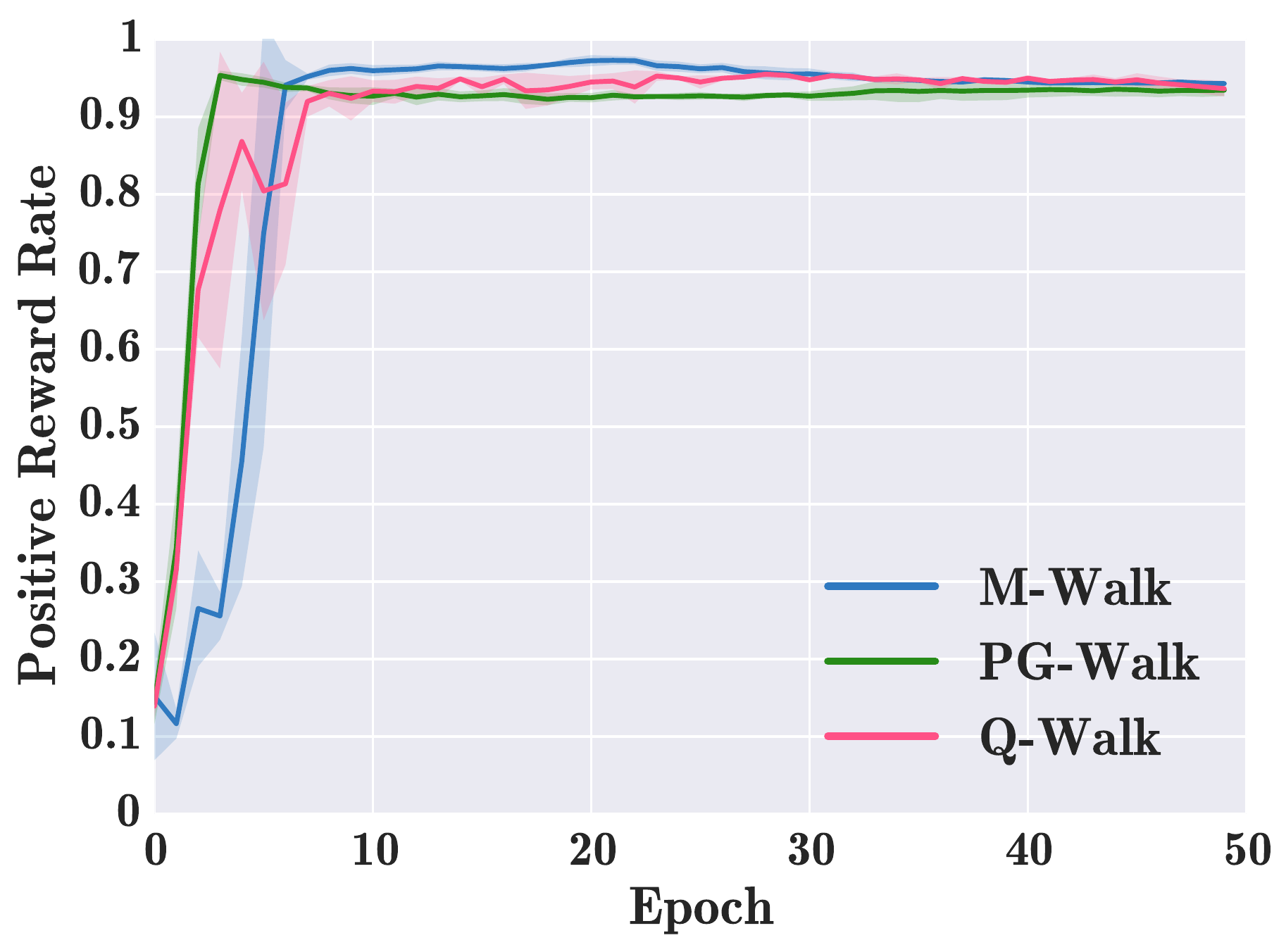}
			}%
		\subfigure[{\scriptsize AthletePlaysSport}]{
			\includegraphics[width=0.31\textwidth]{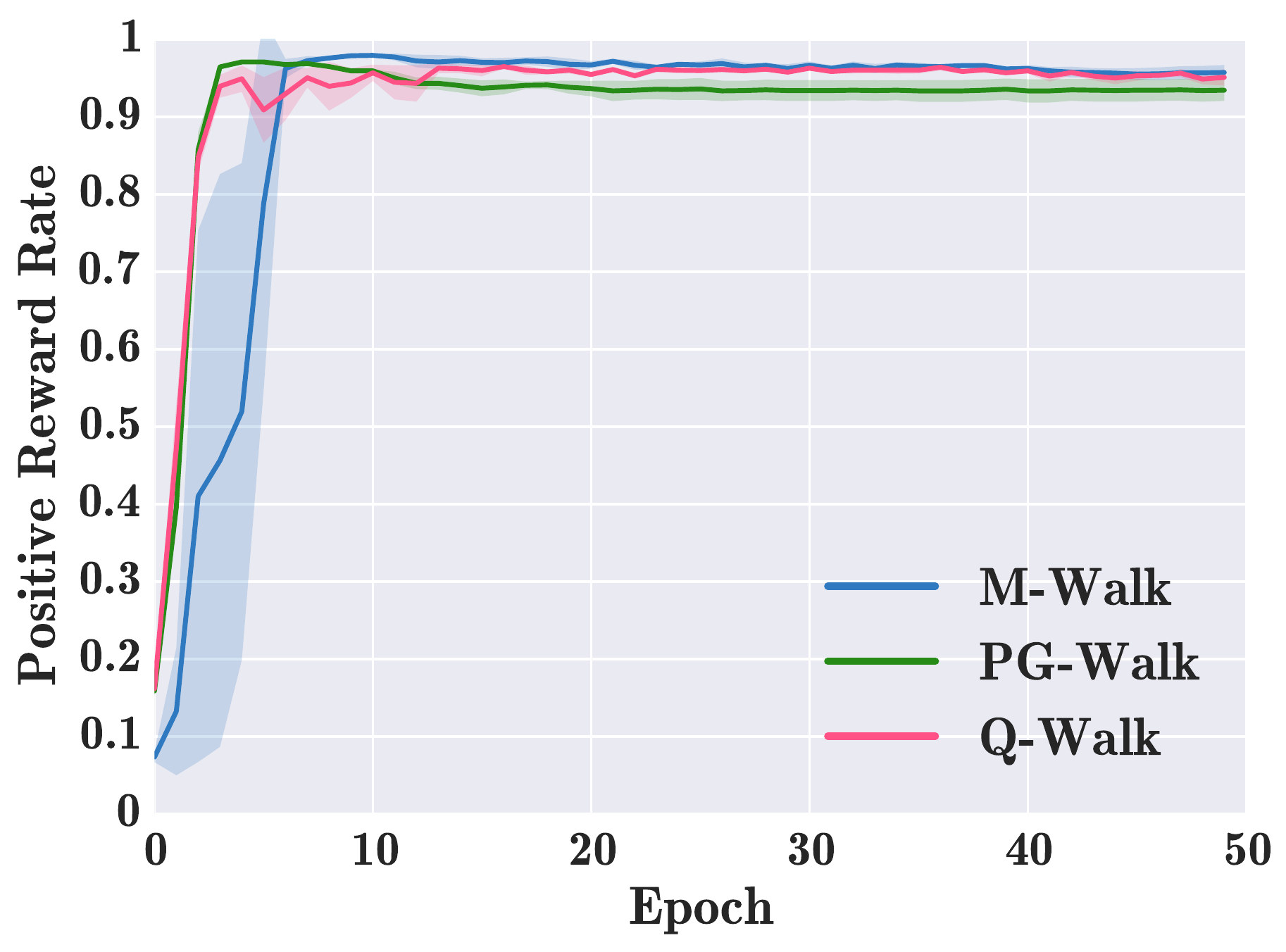}
		} 
			\subfigure[{\scriptsize OrganizationHeadquarterediInCity}]{
			\includegraphics[width=0.31\textwidth]{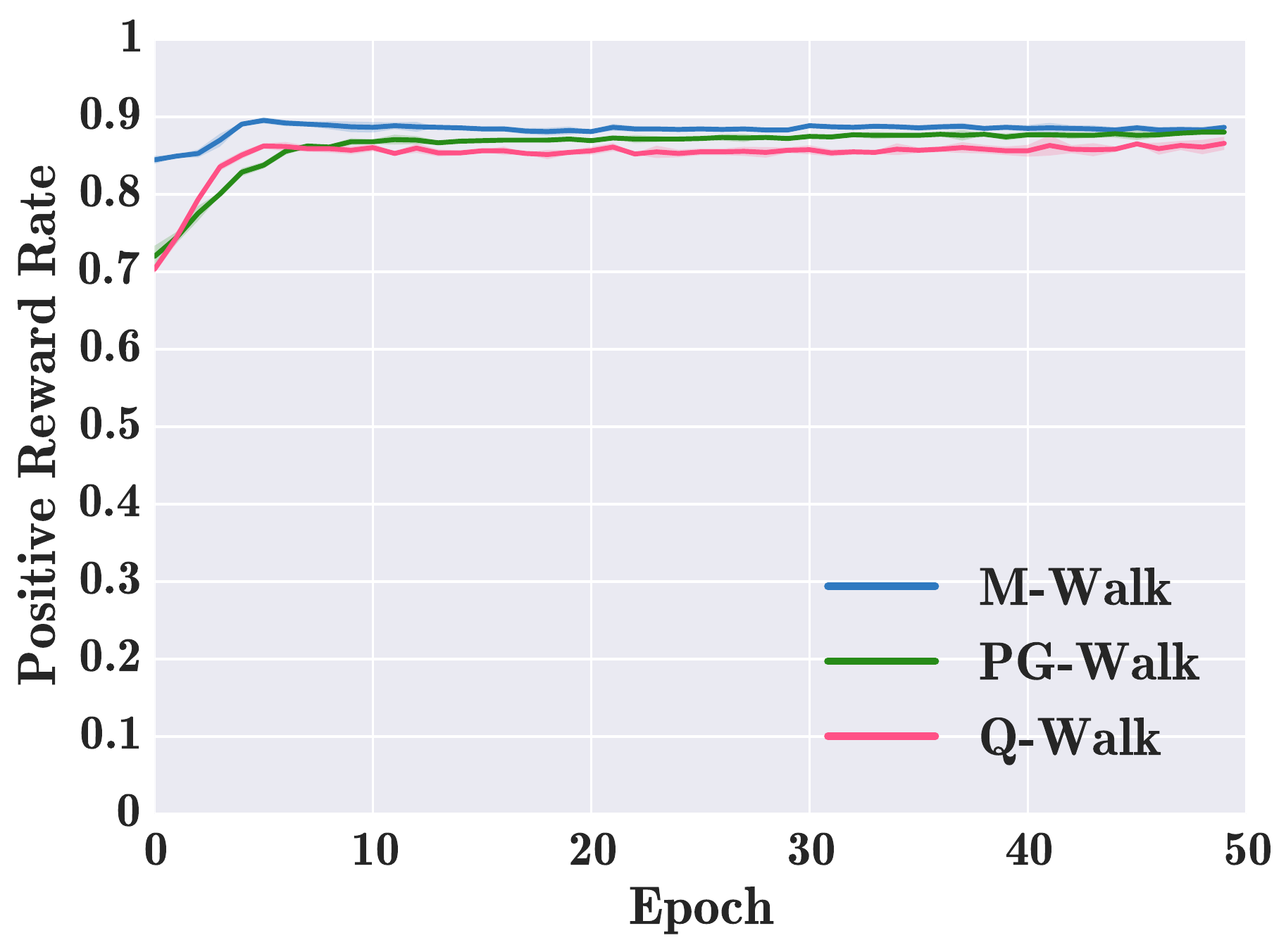}
		}
			\subfigure[\scriptsize PersonBornInLocation]{
			\includegraphics[width=0.31\textwidth]{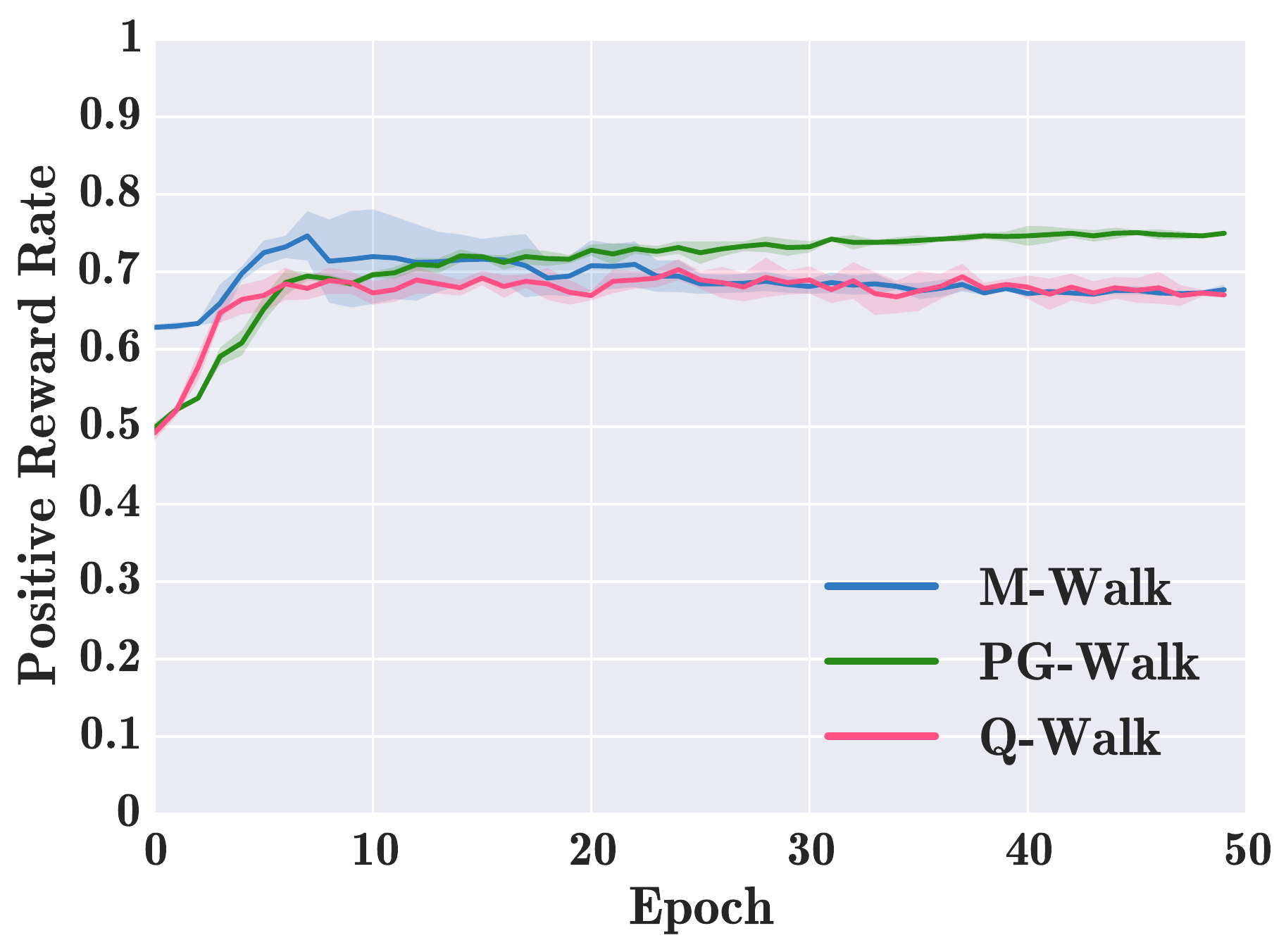}
		}
			\subfigure[\scriptsize AthleteHomeStadium]{
				\includegraphics[width=0.31\textwidth]{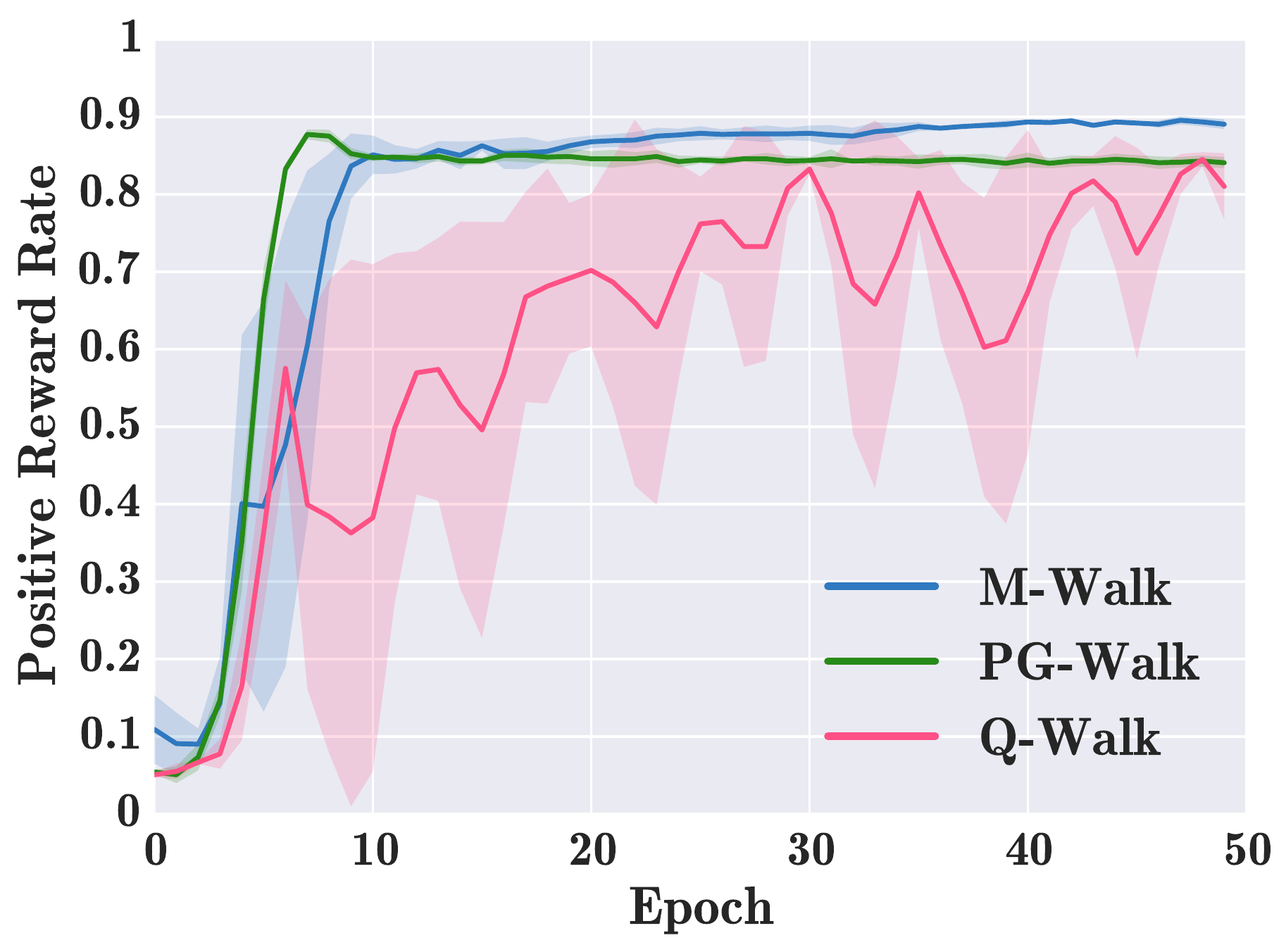}
			}
		\caption{The positive reward rate during training (i.e., percentage of trajectories with positive reward during training) on the NELL-995 task.}
		\label{fig:kbc_train_success_train2}
	\end{figure}

	\begin{figure}[t!]
	\centering
	\subfigure[\scriptsize HITS@3]{
		\label{fig:hyperparameter_hit3_analysis}
		\includegraphics[width=0.31\textwidth]{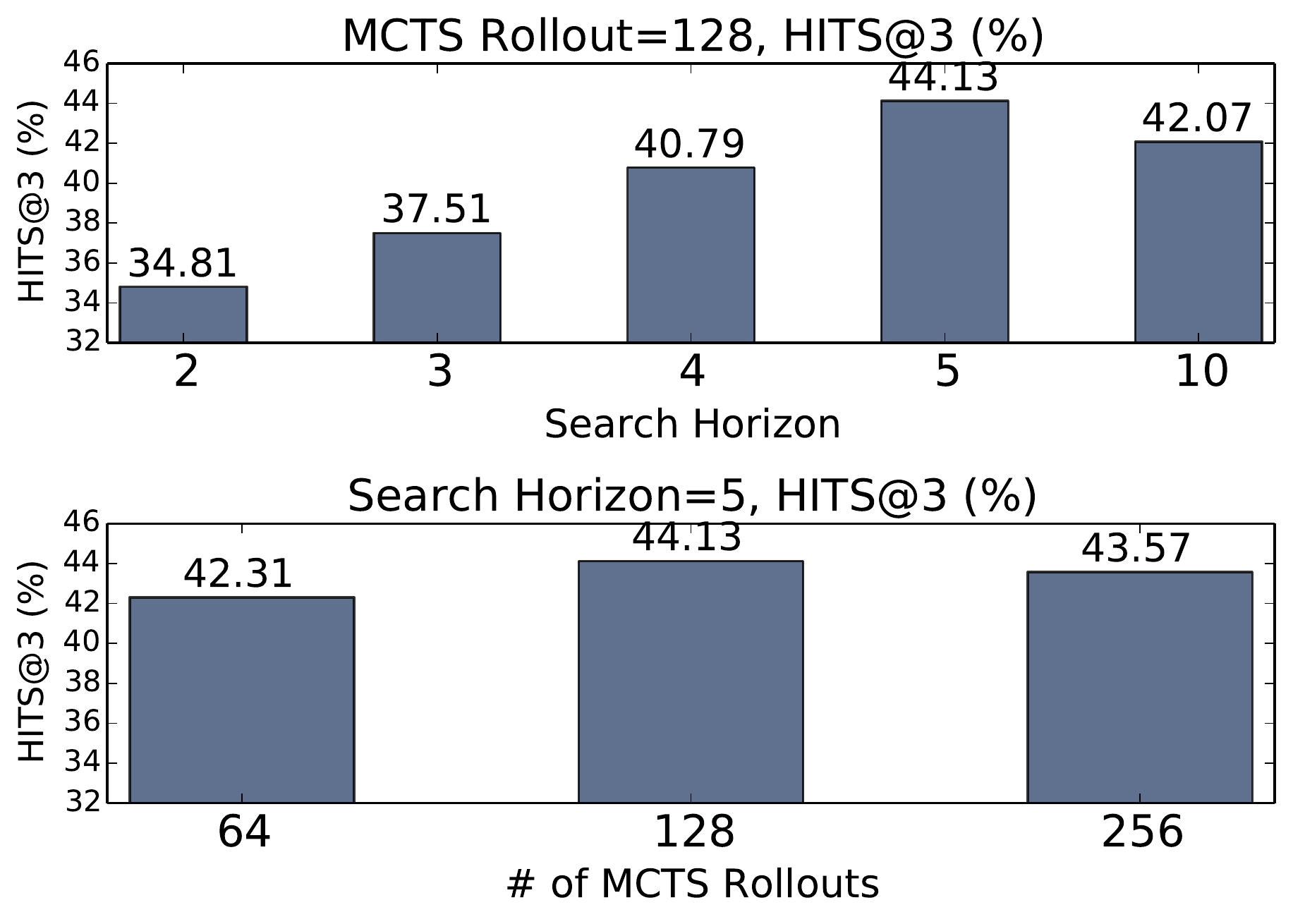}
	}
	\subfigure[\scriptsize HITS@10]{
		\label{fig:hyperparameter_hit10_analysis}
		\includegraphics[width=0.31\textwidth]{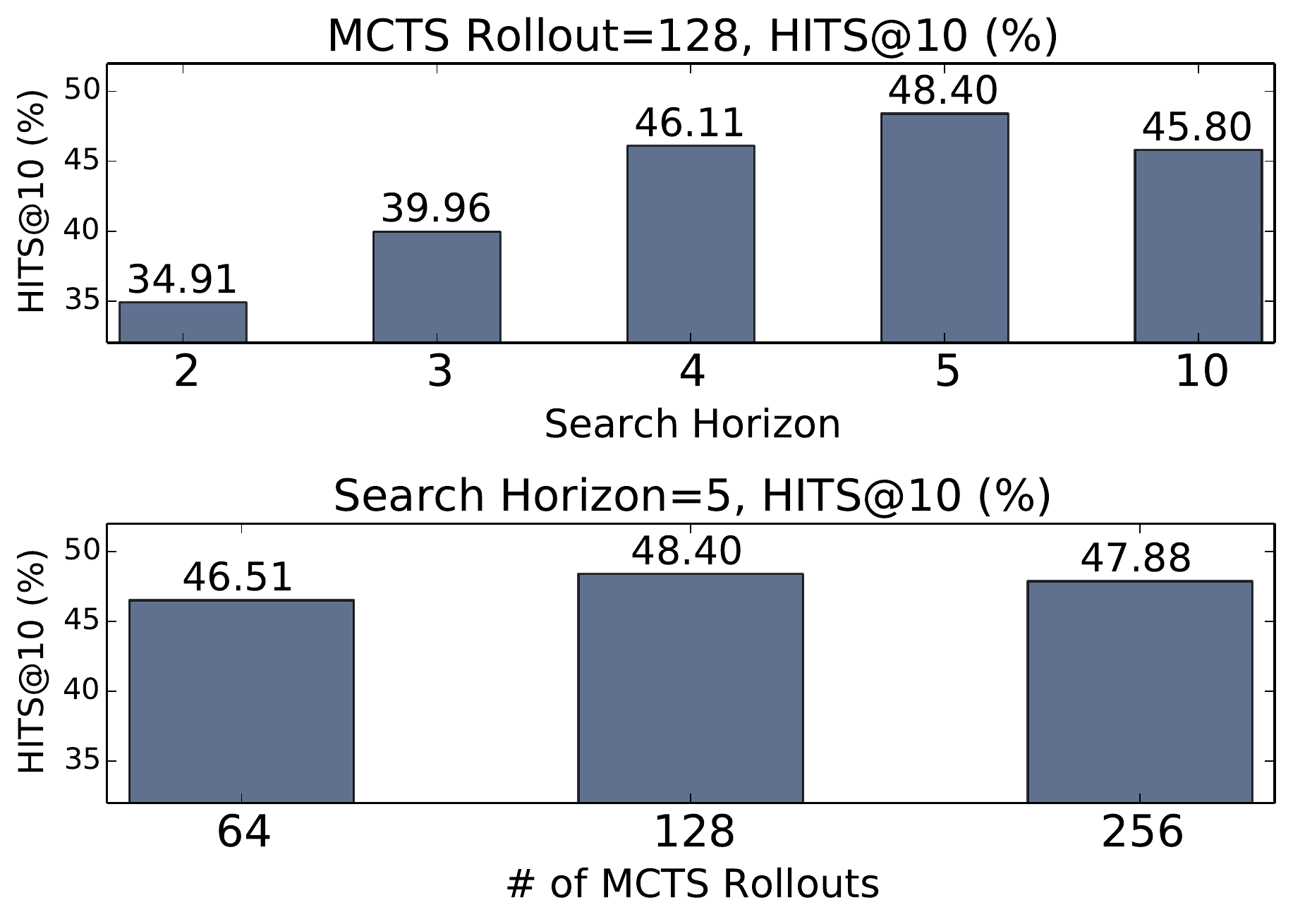}
	}
	\subfigure[\scriptsize Training time]{
		\label{fig:hit1_time}
		\includegraphics[width=0.31\textwidth]{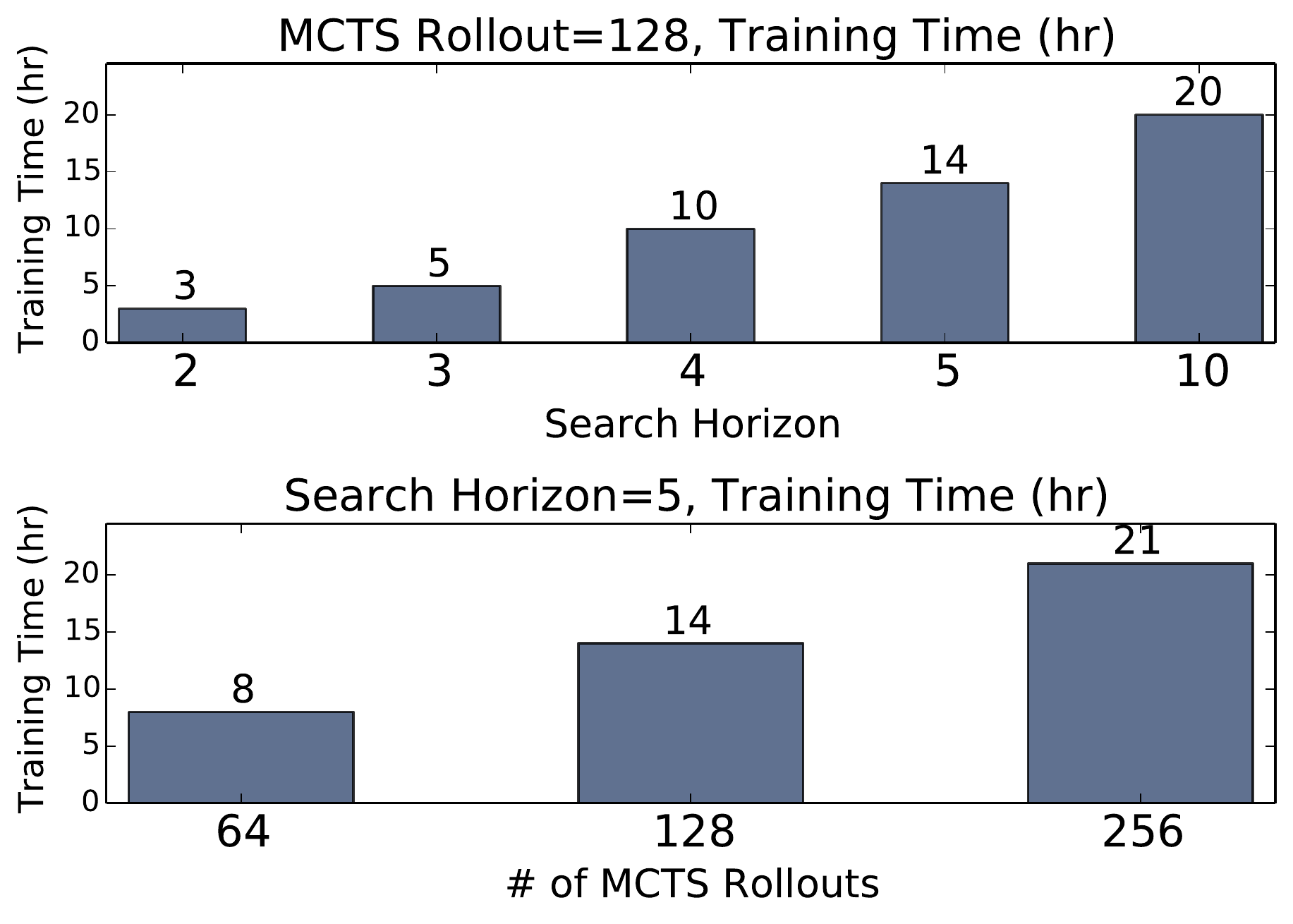}
	}
	\caption{\modelname~hyperparameter and error analysis on WN18RR.}
	\label{fig:wn18rr_analysis2}
	\end{figure}
	
	\begin{table}[t]
    \begin{center}
    \caption{{The HITS@K and MRR results on the NELL995 dataset.}}
    \label{tab:nell995_hits}
    {\scriptsize
		 \resizebox{.98\columnwidth}{!}{%
    \begin{tabular}{lccc|ccccc}
    \hline
    Metric (\%) & \modelname & \multicolumn{1}{c}{PG-Walk} & \multicolumn{1}{c|}{Q-Walk} & MINERVA    & ComplEx    & \multicolumn{1}{l}{ConvE} & \multicolumn{1}{l}{DistMult}   \\\hline
    HITS@1      & {\bf 68.4}  & 66.7 & 66.8 & 66.3 & 61.2 & 67.2 & 61.0  \\
    HITS@3      & {\bf 81.0}  & 77.5 & 77.3 & 77.3 & 76.1 & 80.8 & 73.3  \\
    MRR         & {\bf 75.4}  & 74.8 & 74.5 & 72.5 & 69.4 & 74.7 & 68.0 \\ \hline
    \end{tabular}}
    }%
    \end{center}
    \end{table}

    \begin{table}[ht]
    \begin{center}
    \caption{{The results on the FB15k-237 dataset, in the form of ``mean (standard deviation)''.}}
    \label{tab:fb15k237}
    {\scriptsize
		 \resizebox{.98\columnwidth}{!}{%
    \begin{tabular}{lccc|ccccc}
    \hline
    Metric (\%) & \modelname & \multicolumn{1}{c}{PG-Walk} & \multicolumn{1}{c|}{Q-Walk} & MINERVA    & ComplEx    & \multicolumn{1}{l}{ConvE} & \multicolumn{1}{l}{DistMult} & NeuralLP   \\\hline
    HITS@1      &  16.5(0.3) &  14.8(0.2)   &  15.5(0.2)  & 14.1(0.2) & 20.8(0.2)  & 23.3(0.4) & 20.6(0.4) & 18.2(0.6) \\
    HITS@3      &  24.3(0.2) &  23.3(0.3)   &  23.8(0.4)   & 23.2(0.4) & 32.6(0.5)  & 33.8(0.3) & 31.8(0.2) & 27.2(0.3) \\
    MRR         &  23.2(0.2) &  21.3(0.1)   &  21.8(0.2)  & 20.5(0.3) & 29.6(0.2)  & 30.8(0.2) & 29.0(0.2) & 24.9(0.2) \\ \hline
    \end{tabular}}
    }%
    \end{center}
    \end{table}
	
In this section, we first provide additional experimental results for the NELL995 and WN18RR tasks to support our analysis. In Figure \ref{fig:kbc_train_success_train2}, we show the positive reward rate during training on the NELL995 task. And in Figure \ref{fig:wn18rr_analysis2}, we provide more hyperparameter analysis (search horizon and MCTS simulation number) and training-time analysis. Furthermore, in Table \ref{tab:nell995_hits}, we show the HITS@K and MRR results on NELL995.
	
In addition, we conduct further experiments on the FB15k-237 dataset \cite{Kristinafb15k237}, which is a subset of FB15k \cite{bordes2013translating} with inverse relations being removed. We use the same data split and preprocessing protocol as in \cite{dettmers2018conve} for FB15k-237. The results are reported in Table \ref{tab:fb15k237}. We observe that \modelname~outperforms the other RL-based method (MINERVA). However, it is still worse than the embedding-based methods. In future work, we intend to combine the strength of embedding-based methods and our method to further improve the performance of \modelname.

\subsection{The Reasoning (Traversal) Paths}
\label{Appendix:three_glass_puzzle_paths}
In Table \ref{tab:kbc_example_paths}, we show the reasoning paths of \modelname~on the NELL995 dataset. Each reasoning path is generated by following the edges on the MCTS tree with the highest visiting count $N(s,a)$.

	\begin{table*}[!h]
	\caption{Examples of paths found by \modelname~on the NELL-995 dataset.}
		\centering
 		 \resizebox{\columnwidth}{!}{%
		\begin{tabular}{l l}\hline
		\vspace{1mm}
			(i) \textbf{WorksFor: } 
	        \\
			 \textsf{journalist jerome holtzman} $\xrightarrow{\text{WorksFor}}$? \vspace{-1mm} \\ \\
			 \textsf{journalist jerome holtzman}  $\xrightarrow{\text{JournalistWritesForPublication}}$
			\textsf{website chicago tribune}, (True) 
			\\ \\
			\textsf{politician mufi hannemann} $\xrightarrow{\text{WorksFor}}$? \vspace{-1mm}
			\\ \\
			\textsf{politician mufi hannemann}  $\xrightarrow{\text{PersonHasResidenceInGeopoliticalLocation}}$
			\textsf{city honolulu}, (True) 
			\\ \\
			\textsf{ceo kumar birla} $\xrightarrow{\text{WorksFor}}$? \vspace{-1mm}
            \\ \\
			\textsf{ceo kumar birla}  $\xrightarrow{\text{PersonLeadsOrganization}}$
			\textsf{company hindalco}, (True) 
			\\ \\
			\textsf{professor chad deaton} $\xrightarrow{\text{WorksFor}}$? \vspace{-1mm}
			\\ \\
			\textsf{professor chad deaton}  $\xrightarrow{\text{PersonLeadsOrganization}}$
			\textsf{BiotechCompany baker hughes}, (False) 
			\\ 		
		    \hline\vspace{1mm}
			(ii) \textbf{TeamPlaySport:}
			\\ 
			\textsf{SportsTeam arizona dismond backs} $\xrightarrow{\text{TeamPlaySport}}$? \vspace{-1mm}
			\\ \\
			\textsf{SportsTeam arizona dismond backs}  $\xrightarrow{\text{TeamHomeStadium}}$
			\textsf{StadiumOrEventVenue chase field} $\xrightarrow{\text{SportUsesStadium}^{-1}}$ \textsf{sport baseball}, (True)
			\\ \\
			\textsf{SportsTeam l\_a kings} $\xrightarrow{\text{TeamPlaySport}}$? \vspace{-1mm}
			\\ \\
			\textsf{SportsTeam l\_a kings}  $\xrightarrow{\text{TeamPlaysAgainstTeam}^{-1}}$
			\textsf{SportsTeam red wings} $\xrightarrow{\text{TeamWonTrophy}}$ \textsf{AwardTrophyTournament stanley cup} $\xrightarrow{\text{ChampionshipGameOfTheNationalSport}}$ \textsf{sport hockey}, (True)
			\\ \\
			\textsf{SportsTeam cleveland browns} $\xrightarrow{\text{TeamPlaySport}}$? \vspace{-1mm}
			\\ \\
			\textsf{SportsTeam cleveland browns}  $\xrightarrow{\text{TeamPlaysAgainsTteam}^{-1}}$
			\textsf{SportsTeam yankees} $\xrightarrow{\text{TeamHomeStadium}}$ \textsf{AwardTrophyTournament yankee stadium} $\xrightarrow{\text{SportUsesStadium}^{-1}}$ \textsf{sport baseball}, (False)
			\\\hline
		\end{tabular}
 		}%
		\label{tab:kbc_example_paths}
	\end{table*}

\end{document}